%% file: acl.tex
\pdfoutput=1

\documentclass[11pt]{article}

\usepackage{acl}

\usepackage{times}
\usepackage{latexsym}
\input{math_commands.tex}

\usepackage{bm}
\usepackage{caption}
\usepackage{multirow}
\usepackage{subcaption}
\usepackage{graphicx}
\usepackage{times}
\usepackage{latexsym}
\usepackage{booktabs}
\usepackage[T1]{fontenc}
\usepackage[english]{babel}

\newtheorem{definition}{Definition}

\usepackage[utf8]{inputenc}
\usepackage{xr}
\makeatletter
\newcommand*{\addFileDependency}[1]{
  \typeout{(#1)}
  \@addtofilelist{#1}
  \IfFileExists{#1}{}{\typeout{No file #1.}}
}
\makeatother

\newcommand{\diag}{\mathop{\mathrm{diag}}}

\nolinenumbers

\usepackage{microtype}

\newcommand{\squishlist}{
	\begin{list}{$\bullet$}
		{ \setlength{\itemsep}{0pt}
			\setlength{\parsep}{3pt}
			\setlength{\topsep}{3pt}
			\setlength{\partopsep}{0pt}
			\setlength{\leftmargin}{1.5em}
			\setlength{\labelwidth}{1em}
			\setlength{\labelsep}{0.5em} } }

\newcounter{Lcount}
\newcommand{\squishlisttwo}{
\begin{list}{\arabic{Lcount}. }
	{ \usecounter{Lcount}
		\setlength{\itemsep}{0pt}
		\setlength{\parsep}{0pt}
		\setlength{\topsep}{0pt}
		\setlength{\partopsep}{0pt}
		\setlength{\leftmargin}{2em}
		\setlength{\labelwidth}{1.5em}
		\setlength{\labelsep}{0.5em} } }
		
\newcommand{\squishend}{\end{list} }

\title{Implicit $N$-grams Induced by Recurrence}

\author{Xiaobing Sun \and Wei Lu\\
  StatNLP Research Group\\
  Singapore University of Technology and Design \\
  \texttt{xiaobing\_sun@mymail.sutd.edu.sg, luwei@sutd.edu.sg} \\}
\date{}

\begin{document}
\maketitle
\begin{abstract}
Although self-attention based models such as Transformers have achieved remarkable successes on natural language processing (NLP) tasks, recent studies reveal that they have limitations on \textcolor{black}{modeling sequential transformations} \cite{hahn-2020-theoretical}, which may prompt re-examinations of recurrent neural networks (RNNs) that demonstrated impressive results on handling sequential data.
Despite many prior attempts to interpret RNNs, their internal mechanisms have not been fully understood, and the question on how exactly they capture sequential features remains largely unclear.
In this work, we present a study that shows there actually exist some explainable components
that reside within the hidden states, which are reminiscent of the classical $n$-grams features.
We evaluated such extracted explainable features from trained RNNs on downstream sentiment analysis tasks and found they could be used to model interesting linguistic phenomena such as negation and intensification. 
Furthermore, we examined the efficacy of using such $n$-gram components alone as encoders on \textcolor{black}{ tasks such as sentiment analysis and language modeling}, revealing they could be playing important roles in contributing to the overall performance of RNNs.
We hope our findings could add interpretability to RNN architectures, and also provide inspirations for proposing new architectures for sequential data.
\end{abstract}

\section{Introduction}

Modern recurrent neural networks (RNNs), including Long Short-Term Memory (LSTM) \citep{hochreiter1997long} and Gated Recurrent Units (GRU) \citep{cho-etal-2014-learning}, have demonstrated impressive results on tasks involving sequential data. 
They have proven to be capable of modeling formal languages \citep{weiss-etal-2018-practical, merrill-2019-sequential,merrill-etal-2020-formal2} and \textcolor{black}{capturing structural features \citep{li-etal-2015-hierarchical, li-etal-2015-tree,li-etal-2016-visualizing, linzen-etal-2016-assessing, belinkov-etal-2017-neural, liu-etal-2019-linguistic} on NLP tasks. } 
Although Transformers \citep{NIPS2017_vaswani} have achieved remarkable performances on NLP tasks such as machine translation, it is argued that they may have limitations on modeling hierarchical structure \citep{tran-etal-2018-importance, hahn-2020-theoretical} and cannot handle  functions requiring sequential processing of input well \citep{dehghani2018universal, hao-etal-2019-modeling, bhattamishra-etal-2020-ability, yao-etal-2021-self}. \textcolor{black}{Furthermore, a recent work shows that combining recurrence and attention \citep{lei-2021-attention} can result in strong modeling capacity. Another recent work incorporating recurrent cells into Transformers \citep{recurrent_block_transformer} substantially improved performance on language modeling involving very long sequences, prompting re-investigations of RNNs.} 
On the other hand, it was observed in prior work that RNNs were able to capture linguistic phenomena such as negation and intensification {\color{black}\citep{li-etal-2016-visualizing}}, but the question why they could achieve so still largely remains unanswered.
\begin{figure}
    \centering
    \vspace{-6mm}
    \includegraphics[scale=0.4]{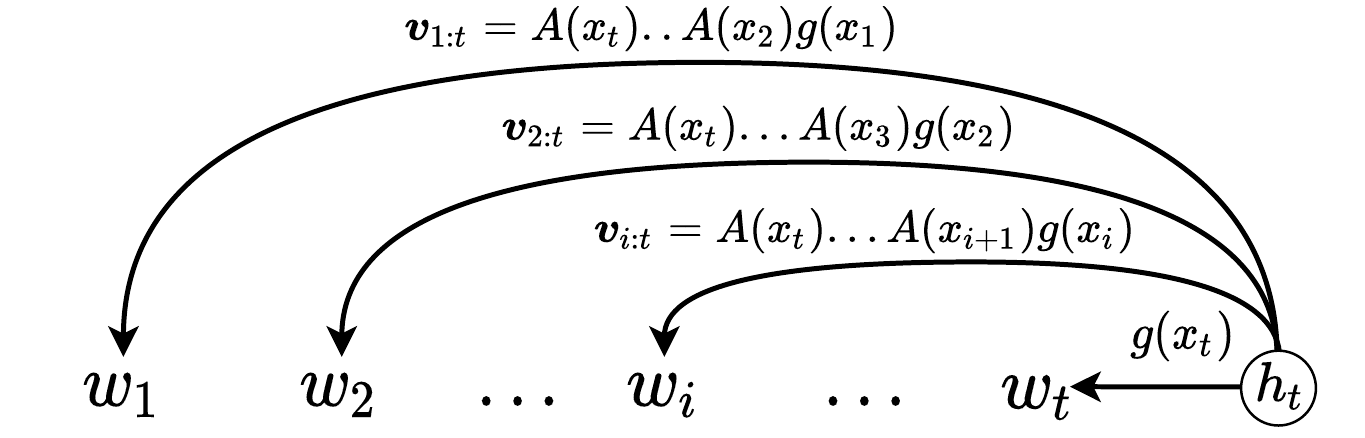}
    \vspace{-3mm}
    \caption{An RNN hidden state may encode a linear combination of all the $n$-grams ending at the current time step.  
    }
    \label{fig:illustration_seq_rep}
    \vspace{-6mm}
\end{figure}

In this work, we {\color{black}focus on better understanding RNNs} from a more theoretical perspective.
We demonstrate that the recurrence mechanism of RNNs may induce a linear combination of interpretable components. These components reside in their hidden states in the form of the iterated matrix-vector multiplication that is based on the representations of tokens in the (reverse) order they appear in the sequence.
Such components, solely depending on inputs and learned parameters, can be conveniently interpreted and are reminiscent of those compositional features used in classical $n$-gram models \citep{10.5555/1214993jurafsky}.
They may also provide us with insights on how RNNs compose semantics from basic linguistic units.
Our analysis further shows that, the hidden state at each time step includes a weighted combination of components that represent all the ``$n$-grams'' ending at that specific position in the sequence \textcolor{black}{as shown in Figure \ref{fig:illustration_seq_rep}.}
We gave specific representations for the $n$-gram components in Elman RNNs \citep{Elman1990FindingSI}, GRUs and LSTMs.

{\color{black}
We investigated the interpretability of those $n$-gram components on trained RNN models, and found they could explain phenomena such as negation and intensification and reflect the overall polarity on downstream sentiment analysis tasks, where such linguistic phenomena are prevalent. 
Our experiments also revealed that the GRU and LSTM models are able to yield better capabilities in modeling such linguistic phenomena than the Elman RNN model, partly attributed to the gating mechanisms they employed which resulted in more expressive $n$-gram components.
{\color{black}We further show that the linear combination of such components yields effective context representations.}
\textcolor{black}{We explored the effectiveness of such $n$-gram components (along with the corresponding context representations) as alternatives to standard RNNs, and found they can generally yield better results than the baseline compositional methods on several tasks, including sentiment analysis, relation classification, named entity recognition, and language modeling.
}

We hope that our work could give inspirations to our community, serving as a useful step towards proposing new architectures for capturing contextual information within sequences.\footnote{Our code is available at {\url{https://github.com/richardsun-voyager/inibr}}.}
}

\section{Related Work}
\paragraph{Interpretability of RNNs:}
A line of work focuses on the relationship between RNNs and finite-state machines \citep{weiss-etal-2018-practical, merrill-2019-sequential, suzgun-etal-2019-lstm, merrill-etal-2020-formal2, DBLP:journals/corr/waf_spectral, pmlr-v89-rabusseau19a},
providing explanation and prediction on the expressive power and limitations of RNNs on formal languages both empirically and theoretically. 
\citet{NIPS2017_6647} investigated conditions that could prevent gradient explosions for GRU based on dynamics. \citet{Maheswaranathan2019ReverseER} and \citet{ pmlr-v119-maheswaranathan20a} linearized the dynamics of RNNs around fixed points of hidden states and elucidated contextual processing. 
Our work focuses on studying a possible mechanism of RNNs that handles exact linguistic features.

Another line of work aims to detect linguistic features captured by RNNs. Visualization approaches \citep{DBLP:journals/corr/KarpathyJL15, li-etal-2016-visualizing} were initially used to examine compositional information in RNN outputs. \citet{linzen-etal-2016-assessing} assessed LSTMs' ability to learn syntactic structure and \citet{pmlr-v139-emami21b} gave rigorous explanations on the standard RNNs' ability to capture long-range dependencies. 
Decomposition methods \citep{murdoch2017automatic,DBLP:journals/corr/abs-1801-05453murdoch,singh2018hierarchical_acd, arras-etal-2017-explaining,arras2019explaining,chen-etal-2020-generating} were proposed to produce importance scores for hierarchical interactions in RNN outputs. 
Our work can be viewed as an investigation on how those interaction came about.


\paragraph{Compositional Models:}
A variety of compositional functions based on vector spaces have been proposed in the literature to compose semantic meanings of phrases, including simple compositions of adjective-noun phrases represented as matrix-vector multiplication \citep{mitchell-lapata-2008-vector, baroni-zamparelli-2010-nouns}
and a matrix-space model \citep{rudolph-giesbrecht-2010-compositional, yessenalina-cardie-2011-compositional} based on matrix multiplication.
\citet{socher-etal-2012-semantic, socher2013recursive} introduced a recursive neural network model that assigns every word and longer phrase in a parse tree both a vector and a matrix, and represents composition of a non-terminal node with matrix-vector multiplication.
\citet{kalchbrenner-blunsom-2013-recurrent-convolutional} employed convolutional and recurrent neural networks to model compositionality at the sentence and discourse levels respectively. 
Those models are designed in an intuitive manner based on the nature of languages thus being interpretable.
{\color{black}We can show that RNNs may process contextual information in a way bearing a resemblance to those early models.}

\section{A Theory on $N$-gram Representation}


\begin{table*}[h!]
\centering
\scalebox{0.6}{
\begin{tabular}{llrcl}
\toprule
\multirow{2}{*}{ \textbf{Model}}  & \multicolumn{1}{l}{\bf $N$-gram}                                               & \multicolumn{1}{r}{\bf Context }&\multirow{2}{*}{$L$}&\multirow{2}{*}{ \textbf{Representative Work}}  \\
&{\bf Representation}&{\bf Representation}&\\
\midrule
Vector Multiplicative
& \multirow{2}{*}{$\vv_{i:j}=g(x_i)\odot  \dots \odot g(x_j){\color{white}\left(\prod_{k=t}^{i+1} A(x_k)\right)}$\!\!\!\!\!\!\!\!\!\!\!\!\!\!\!\!\!\!\!\!\!\!\!\!\!\!\!\!\!\!\!\!\!\!\!\!\!}         
&\multirow{2}{*}{$\vv_{1:t}$}  &\multirow{2}{*}{$t$}
&      \multirow{2}{*}{\small\citet{mitchell-lapata-2008-vector}}    
\\
\ ({\bf\textsc{vm}})
\\
Matrix Multiplicative
& \multirow{2}{*}{$\mM_{i:j}=\prod_{k=i}^j A(x_k){\color{white}\left(\prod_{k=t}^{i+1} A(x_k)\right)}\!\!\!\!\!\!\!\!\!\!\!\!\!\!\!\!\!\!\!\!\!\!\!$}
&\multirow{2}{*}{$\mM_{1:t}$}&\multirow{2}{*}{$t$}
&     \multirow{2}{*}{\small \citet{yessenalina-cardie-2011-compositional}}
\\
\ ({\bf\textsc{mm}})
\\
{Vector Additive ({\em weighted})}
&
\multirow{2}{*}{$\vv_{i:j}=\mC_{j-i} g(x_i){\color{white}\left(\prod_{k=t}^{i+1} A(x_k)\right)}$\!\!\!\!\!\!\!\!\!\!\!\!\!\!\!\!\!\!\!\!\!\!\!\!}
&
\multirow{2}{*}{$\!\!\!\!\!\!\!\!\!\!\!\!\sum_{i=t-m+1}^{t} \vv_{i:t}$} &\multirow{2}{*}{$m$}
&
\multirow{2}{*}{\small \citet{bengio2003neural}}                         \\
\ ({\bf \textsc{va-w}})
\\
Vector Additive ({\em exponentially weighted})                                                        
&
\multirow{2}{*}{$\vv_{i:j}=\mC^{j-i}g(x_i){\color{white}\left(\prod_{k=t}^{i+1} A(x_k)\right)}\!\!\!\!\!\!\!\!\!\!\!\!\!\!\!\!\!$}         &
\multirow{2}{*}{$\sum_{i=1}^{t} \vv_{i:t}$} 
&
\multirow{2}{*}{$t$}
&
\multirow{2}{*}{\small \citet{pmlr-v139-emami21b}}
\\
\ ({\bf\textsc{va-ew}})
\\
Matrix-Vector Multiplicative ({\em restricted})                                                   & \multirow{2}{*}{$\vv_{i-1:i}=A(x_{i-1})g(x_i){\color{white}\left(\prod_{j=t}^{i+1} A(x_j)\right) g(x_i)}$\!\!\!\!\!\!\!\!\!\!\!\!\!\!\!\!\!\!\!\!\!\!\!\!\!\!\!\!\!\!\!\!\!\!\!\!\!\!\!\!\!\!\!\!\!\!\!\!\!\!\!\!\!\!\!\!}                             &\multirow{2}{*}{$\vv_{t-1:t}$}&\multirow{2}{*}{$2$}&    \multirow{2}{*}{\small \citet{baroni-zamparelli-2010-nouns}}                      \\
\ ({\bf\textsc{mvm-r}})\\
 Matrix-Vector Multiplicative                                                  & \multirow{2}{*}{$\vv_{i:j}=\left(\prod_{k=j}^{i+1} A(x_k)\right) g(x_i)$}                              &\multirow{2}{*}{$ {\vv_{1:t}}$} &\multirow{2}{*}{$t$}&    \multirow{2}{*}{-}                            \\
\ ({\bf\textsc{mvm}})\\
\midrule
{Matrix-Vector Multiplicative-Additive}                                    & \multirow{2}{*}{$\vv_{i:j}=\left(\prod_{k=j}^{i+1} A(x_k)\right) g(x_i)$}  &\multirow{2}{*}{$\sum_{i=1}^t {\vv_{i:t}}$} &\multirow{2}{*}{$t$}& \multirow{2}{*}{This work}
\\
\ ({\bf\textsc{mvma}})\\
\bottomrule
\end{tabular}
}
\vspace{-2mm}
\caption{Different models for defining representations for $n$-grams within the phrase $x_1,x_2,\dots ,x_{t-1},x_t$ and constructing the context representation out of the $n$-grams during learning. $L$: the maximum length allowed for the context representation. $\mC$ is a weight matrix, and $\mC_{k}$ is a (relative) position-specific weight matrix. $A$ and $g$ are functions that return a matrix and a vector respectively. 
}
\vspace{-4mm}
\label{tab:compositional_functions_comparison}
\end{table*}

First, let us spend some time to discuss how to represent $n$-grams.
Various approaches to representing $n$-grams have been proposed in the literature \cite{mitchell-lapata-2008-vector,bengio2003neural, mitchell-lapata-2008-vector, 10.5555/mnih,10.2307/linear_rnn, Orhan2020Improved, pmlr-v139-emami21b,rudolph-giesbrecht-2010-compositional,yessenalina-cardie-2011-compositional,baroni-zamparelli-2010-nouns}. 
We summarize in Table \ref{tab:compositional_functions_comparison} different approaches for representing $n$-grams.

Although empirically it has been shown that different approaches can lead to different levels of effectiveness,
the rationales underlying many of the design choices remain unclear.
In this section, we establish a small theory on representing $n$-grams, which leads to a new formulation on capturing the semantic information within $n$-grams.

Let us assume we have a vocabulary $\sV$ that consists of all possible word tokens.
The set of $n$-grams can be denoted as $\sV^*$ (including the special $n$-gram which is the empty string $\epsilon$).
Consider three $n$-grams $a$, $b$, and $c$ from $\sV^*$, with their semantic representations $r(a)$, $r(b)$, and $r(c)$ respectively.
Similarly, we may have $r(ab)$ which return the semantic representations of the concatenated $n$-grams $ab$.
It is desirable for our representations to be compositional in some sense. Specifically, a longer $n$-gram may be semantically related to those shorter $n$-grams it contains in some way.

Under some mild compositional assumptions related to the {\em principle of compositionality} \cite{frege1948sense}\footnote{The principle states that ``the meaning of an expression is determined by the meanings of the sub-expressions it contains and the rules used to combine such sub-expressions''.}, it is reasonable to expect that there exists some sort of rule or operation that allows us to compose semantics of longer $n$-grams out of shorter ones.
Let us use $\otimes$ to denote such an operation.
We believe a good representation system for $n$-grams shall satisfy several key properties.
First, the semantics of the $n$-gram $abc$ shall be determined through either combining the semantics of the two $n$-grams $a$ and $bc$ or through combining the semantics of $ab$ and $c$.
The semantics of $abc$ is unique, regardless of which of these two ways we use.
Second, for the empty string $\epsilon$, it should not convey any semantics.
Formally, we can write them  as:\footnote{Besides, another important property is that the order used for combining two $n$-grams does matter.
In other words,  $r(a)\otimes r(b)$ usually may not be the same as $r(b)\otimes r(a)$.}
\squishlist
\item {\bf Associativity}: $\forall a,b,c\in\sV^*$, $ (r(a)\otimes r(b))\otimes r(c) = r(a)\otimes (r(b)\otimes r(c))$
\item {\bf Identity}: $\forall a\in\sV^*$, $r(a)\otimes r(\epsilon) =r(a)$, and $r(\epsilon)\otimes r(a) = r(a)$
\squishend

This essentially shows that the representation space for all $n$-grams under the operation $\otimes$, denoted as $(\sV^*, \otimes)$, forms a {\em monoid}, an important concept in abstract algebra \cite{lallement1979semigroups}, with significance in theoretical computer science \cite{meseguer1990petri,rozenberg2012handbook}.

On the other hand, it can be easily verified that the space of all $d\times d$ (where $d$ is an integer) real square matrices under matrix multiplication, denoted as $(\sR^{d\times d},\cdot)$, also strictly forms a monoid (i.e., it is associative  and has an identity, but is {\em not} commutative).
We can therefore establish a homomorphism from $\sV^*$ to $\sR^{d\times d}$, resulting in the function $r(\cdot)\in \sV^*\rightarrow \sR^{d\times d}$.

This essentially means that we may be able to rely on a sub-space within $\sR^{d\times d}$ as our mathematical object to represent the space of $n$-grams, where the matrix multiplication operation can be used to compose representations for longer $n$-grams from shorter ones.
Thus, for a unigram $x$ (a single word in the vocabulary), we have:
\begin{equation}
\vspace{-2mm}
\small
r(x)
:=
\mA_x
\end{equation}
where $\mA_x\in\sR^{d\times d}$ is the representation for the word $x$ (how to learn such a matrix is a separate question to be discussed later).
Note that the empty string $\epsilon$ comes with a unique representation which is the ${d\times d}$ identity matrix $\mI$.

We can either use matrix left-multiplication or right-multiplication as our operator $\otimes$.
Assume the language under consideration employs the left-to-right writing system.
It is reasonable to believe that a human reader processes the text left-to-right, and the semantics of the text gets evolved each time the reader sees a new word.
We may use the matrix left-multiplication as the preferred operator in this case.
{\color{black}The system will left-multiply (modify) an existing $n$-gram representation with a matrix associated with the new word that appears right after the existing $n$-gram, forming the representation of the new $n$-gram.}
Such an operation essentially performs a transform that simulates the process of yielding new semantics when appending a new word at the end of an existing phrase.
With this, for a general $n$-gram $x_i,x_{i+1},\dots,x_t$ ($i\leq t$), we have:
\begin{equation}
\vspace{-2mm}
\small
r(x_i, x_{i+1}, \dots, x_t)
=
\prod_{k=t}^i \mA_{x_k}
\end{equation}

However, the conventional wisdom in NLP has been to use vectors to represent basic linguistic units such as words, phrases or sentences \cite{mikolov2013efficient,mikolov2013distributed,pennington2014glove,kiros2015skip}.
This can be achieved by a transform:
\begin{equation}
\vspace{-2mm}
\label{eqn:ngram_u}
\small
\left(\prod_{k=t}^i \mA_{x_k}\right)
\vu
\end{equation}
where $\vu\in\sR^d$ is a  vector that maps the resulting matrix representation into a vector representation.

Next, we will embark on our  journey to examine the internal representations of RNNs.
As we will see, interestingly, our developed $n$-gram representations can emerge within such models.

\section{Interpretable Components in RNNs}

An RNN is a parameterized function whose hidden state can be written recursively as:
\vspace{-1mm}
\begin{equation}
\vspace{-1mm}
\small
    \begin{aligned}
       \vh_t = f(x_t, \vh_{t-1}),
    \end{aligned}
    \label{eq:general_rnn_def}
\end{equation}
where $x_t$ is the input token at time step $t$ and $\vh_{t-1} \in \sR^d$ is the previous hidden state.
Assume $f$ is differentiable at any point, with the Taylor expansion, $\vh_t$ can be rewritten as:
\vspace{-0mm}
\begin{equation}
\vspace{-2mm}
\small
    \begin{aligned}
       \vh_t = f(x_t, \boldsymbol{0}) + \nabla f (x_t, \boldsymbol{0}) \vh_{t-1}  +   o(\vh_{t-1}),
    \end{aligned}
\end{equation}
where $\nabla  f (x_t, \boldsymbol{0}) = \frac{\partial f}{\partial \vh_{t-1}} |_{\vh_{t-1}=\boldsymbol{0}}$ is the Jacobian matrix, and $o$ is the remainder of the Taylor series. 

\textcolor{black}{Let  $g(x_t) = f(x_t, \boldsymbol{0})$ and $A(x_t) = \nabla f (x_t, \boldsymbol{0})$.
Note that $g(x_t) \in \sR^d$ and $A(x_t) \in \sR^{d \times d}$ are both functions of $x_t$. Therefore, the equation above can be written as:
\vspace{-0mm}
\begin{equation}
\vspace{-2mm}
\small
    \begin{aligned}
       \vh_t = g(x_t) + A(x_t) \vh_{t-1} +  o(\vh_{t-1}).
    \end{aligned}
    \label{eq:rnn_taylor_expansion}
\end{equation}
}

If the hidden state has a sufficiently small norm, it can be approximated by the first-order Taylor expansion as follows\footnote{There will be an ``approximation gap'' at each time step between the ``approximated'' hidden state and the actual standard hidden state.
We may leverage regularization methods such as weight-decaying and the spectral normalization \citep{miyato2018spectral} to prevent the gap from growing unbounded.}:
\vspace{-0mm}
\begin{equation}
\small
    \begin{aligned}
\vspace{-2mm}
       \vh_t \approx g(x_t) + A(x_t) \vh_{t-1}.
    \end{aligned}
    \label{approx:hidden_recurrence_relation}
\end{equation}

Next we illustrate how this recurrence relation can help us identify some salient components.

\subsection{Emergence of $N$-grams}

Consider the simplified RNN with the following recurrence relation,
\begin{equation}
\small
    \begin{aligned}
       \vh_t = g(x_t) + A(x_t) \vh_{t-1},
    \end{aligned}
    \label{eqn:linear_recurrent_rep}
\end{equation}
where $\vh\in\sR^d$, and $g(x_t)\in \sR^d$ and $A(x_t) \in \sR^{d \times d}$ are functions of $x_t$.
This recurrence relation can be expanded repeatedly as follows,
\begin{equation}
\small
    \begin{aligned}
    \label{eq:linear_recurrent_expand}
       \vh_t 
       &\! =\! g(x_t) \!+\! A(x_t) g(x_{t-1}) \!+\! A(x_t) A(x_{t-1}) \vh_{t-2} \nonumber \\ 
       &\!= \dots = \sum_{i=1}^{t} A(x_{t})\dots A(x_{i+1}) g(x_i)\nonumber\\ 
       &\!=\!\sum_{i=1}^{t}\underbrace{\left(\prod^{i+1}_{j=t} A(x_j) \right)g(x_i)}_{\vv_{i:t}},
    \end{aligned}
\end{equation}

We can see that $\vv_{i:t}$ bear some resemblance to the term in Equation \ref{eqn:ngram_u}, which can be rewritten as:
\vspace{-2mm}
\begin{equation}
\vspace{-3mm}
\small
    \label{eqn:specific_representation}
    \begin{aligned}
      \left(\prod_{j=t}^{i+1} \underbrace{\mA_{x_j}}_{A(x_j)}\right) \underbrace{\Big(\mA_{x_{i}}\vu\Big)}_{g(x_i)},
    \end{aligned}
\end{equation}

With the definition $A(x_j):=\mA_{x_j}$ and $g(x_i):=A(x_{i})\vu$, we can see $\vv_{i:t}$ can be interpreted as an ``$n$-gram representation'' that we developed in the previous section.
It is important to note that, however, the use of function $g(x_i)$  in RNNs may lead to greater expressive power than the original formulation based on $\mA_{x_{i}}\vu$.\footnote{This is because we can always construct $g(x_i)$ from any given $\mA_{x_i}$ and $\vu$, but in general we may not always be able to decompose $g(x_i)$ into the form $\mA_{x_i}\vu$ (for all $x_i$).}

This interesting result shows that the hidden state of a simple RNN (characterized by Equation \ref{eqn:linear_recurrent_rep}) is the sum of the representations of all the $n$-grams ending at time step $t$.
Such salient components within RNN also show that the standard RNN may actually have a mechanism that is able to capture implicit $n$-gram information as described above. This leads to the following definition:
\begin{definition}[$N$-gram Representation]
\label{def:specific_representation}
For the $n$-gram $x_i, x_{i+1}, \dots , x_t$,  its representation is:
\vspace{-2mm}
\begin{equation}
\small
\vspace{-2mm}
    \label{eqn:specific_representation}
    \begin{aligned}
      \vv_{i:t}&= \left(\prod_{j=t}^{i+1} A(x_j)\right) g(x_i),
    \end{aligned}
    \vspace{-1mm}
\end{equation}
where ${A}(x_j) \in \sR^{d \times d}$ and $g(x_i)\in\sR^d$.
\end{definition}

\subsection{Context Representation}

With the above definition, we may want to consider how to perform learning.
The learning task involves identifying the functions $A$ and $g$ -- in other words, learning representations for word tokens.

A typical learning setup that we may consider here is the task of language modeling.
Such a task can be defined as predicting the next word $x_{t+1}$ based on the representation of preceding words $x_1,x_2,\dots,x_t$ which serves as its left context.
This is an unsupervised learning task, where the underlying assumption involved is the {\em distributional hypothesis} \cite{harris1954distributional}.
Specifically, the model learns how to ``reconstruct'' the current word $x_{t+1}$ out of $x_1,x_2,\dots,x_t$ which serves as its context.

Now the research question  is how to define the representation for this specific context.
As this left context is also an $n$-gram, it might be tempting to directly use its $n$-gram representation defined above to characterize such a left context.
However, we show such an approach is not desirable.

The $n$-gram representation for this context can be written in the following alternative form:
\vspace{-2mm}
\begin{equation}
\vspace{-2mm}
\small
    \label{eqn:hidden_rep}
    \begin{aligned}
\vspace{-2mm}
      \vv_{1:t}&= \left(\prod_{j=t}^{2}\! A(x_j)\right) \!g(x_i)=W(x_{2:t}) g(x_1),
    \end{aligned}
\end{equation}

This shows that the $n$-gram representation of $x_1,x_{2},\dots,x_t$ could be interpreted as a ``weighted'' representation of the  word $x_1$ (where the weight matrix is derived from the words between $x_1$ and $x_{t+1}$, measuring the strength of the connection between them).
However, ideally, the context representation shall not just take $x_1$ but other adjacent words preceding $x_{t+1}$ into account, where each word contributes towards the final context representation based on the connection between them. This leads to the following way of defining the context:
\vspace{-2mm}
\begin{equation}
\vspace{-2mm}
\small
    \begin{aligned}
     \sum_{i=1}^t \vv_{i:t}= \sum_{i=1}^{t}\left(\prod_{j=t}^{i+1} A(x_j)\right) g(x_i)
     \\
     = \sum_{i=1}^{t} W(x_{i:t}) g(x_i),
    \end{aligned}
\end{equation}

In fact, such an idea of defining the context as a weighted combination of surrounding words is not new -- it recurs in the literature of language modeling \cite{bengio2003neural,10.5555/mnih}, word embedding learning \cite{mikolov2013efficient,mikolov2013distributed}, and graph representation learning \cite{cao2016deep}.

Interestingly, the hidden states in the RNNs, as shown in Equation \ref{eq:linear_recurrent_expand}, also suggest exactly the same way of defining this left context.
Indeed, when using RNNs for language modeling, each hidden state is exactly serving as the context representation for predicting the next word in the sequence.

The above gives rise to the following definition:
\begin{definition}[Context Representation]
\label{def:specific_representation}
For the $n$-gram $x_1, x_2, \dots , x_t$,  its  representation when serving as the (left) context is:
\vspace{-2mm}
\begin{equation}
\vspace{-2mm}
\small
    \begin{aligned}
      {{\vc}}_{1:t}&= \sum_{i=1}^t \vv_{i:t}= \sum_{i=1}^{t}\left(\prod_{j=t}^{i+1} A(x_j)\right) g(x_i),
    \end{aligned}
\end{equation}
where ${A}(x_j) \in \sR^{d \times d}$ and $g(x_i)\in\sR^d$.
\end{definition}

\begin{table*}[t!]
\vspace{-4mm}
\centering
\scalebox{0.55}{
\begin{tabular}{lllllllllllllll}
\toprule
                      & \multicolumn{4}{c}{\bf Definition}        & \multicolumn{10}{c}{\bf Parameterization}                 \\
                      \midrule
\multirow{2}{*}{\rotatebox[origin=c]{90}{\small Elman}} & \multicolumn{4}{l}{{\multirow{2}{*}{
$\vh_t = \tanh(\mW_{in}\vx_t \!+\! \mW_{ih} \vh_{t-1})$
}}
}                  & \multicolumn{5}{l}{{\multirow{2}{*}{
$A(x_t)=\diag [\tanh'(\mW_{in}\vx_t)] \mW_{ih}$
}}
}                  & \multicolumn{5}{l}{{\multirow{2}{*}{
 \ \ \ \ \ $g(x_t) = \tanh(\mW_{in}\vx_t)$.
 }}
}                  \\
                       & \multicolumn{4}{l}{}                  & \multicolumn{5}{l}{}                  & \multicolumn{5}{l}{}                  \\
 \midrule                      
                       
\multirow{5}{*}{\rotatebox[origin=c]{90}{\small GRU}} & \multicolumn{4}{l}{{\multirow{5}{*}{
\begin{tabular}[l]{@{}rl@{}}
$\vr_t $\!\!\!\!\!&$= \sigma (\mW_{ir} \vx_t + \mW_{hr} \vh_{t-1})$\\
$\vz_t $\!\!\!\!\!&$= \sigma (\mW_{iz} \vx_t + \mW_{hz} \vh_{t-1})$ \\ 
$\vn_t $\!\!\!\!\!&$= \tanh (\mW_{in} \vx_t \!+\! \vr_t \! \odot\! \mW_{hn} \vh_{t-1})$ \\ 
$\vh_t $\!\!\!\!\!&$= (1-\vz_t) \odot \vn_t + \vz_t \odot \vh_{t-1}$  
\end{tabular}
}}
}                 & \multicolumn{5}{l}{{\multirow{5}{*}{
\begin{tabular}[l]{@{}rl@{}}
 $A(x_t) $\!\!\!\!\!&$= \!\diag\left[f_n(x_t) \!\odot\! [1\!-\!g_z(x_t)]\! \odot\! g_r(x_t) \right] \! \mW_{hn} $ \\ 
 \textcolor{white}{$A(x_t)$} \!\!\!\!\!&$-\diag[g_n(x_t) \odot f_z(x_t)]\mW_{hz}$ \\ 
 \textcolor{white}{$A(x_t)$} \!\!\!\!\!&$+\diag[g_z(x_t)]$\\
 $\ \ g(x_t)$\!\!\!\!\!&$= [1-g_z(x_t)]\odot g_n(x_t)$
\end{tabular}
}}
}                  & \multicolumn{5}{l}{{\multirow{5}{*}{
\begin{tabular}[l]{@{}rl@{}}
    where:\\
          $g_r(x_t)  $\!\!\!\!\!&$= \sigma (\mW_{ir} \vx_t),\ \ \ \ \ \ \ \ \ \  f_r(t) = g'_r(x_t)$,\\
      $g_z(x_t)  $\!\!\!\!\!&$= \sigma (\mW_{iz} \vx_t)$,
      \ \ \ \ \ \ \ \ \!$f_z(x_t) = g'_z(x_t)$,\\
      $g_n(x_t)  $\!\!\!\!\!&$= \tanh (\mW_{in} \vx_t)$,
      \ $f_n(x_t) = g'_n(x_t)$.
\end{tabular}
 }}
}                    \\
                       & \multicolumn{4}{l}{}                  & \multicolumn{5}{l}{}                  & \multicolumn{5}{l}{}                  \\
                       & \multicolumn{4}{l}{}                  & \multicolumn{5}{l}{}                  & \multicolumn{5}{l}{}                  \\
                        & \multicolumn{4}{l}{}                  & \multicolumn{5}{l}{}                  & \multicolumn{5}{l}{}                  \\
                       & \multicolumn{4}{l}{}                  & \multicolumn{5}{l}{}                  & \multicolumn{5}{l}{}                  \\

\midrule
\multirow{9}{*}{\rotatebox[origin=c]{90}{\small LSTM}} & \multicolumn{4}{l}{\multirow{9}{*}{
\begin{tabular}[l]{@{}rl@{}}
   $\vi_t $\!\!\!\!\!&$= \sigma (\mW_{ii} \vx_t + \mW_{hi} \vh_{t-1})$ \\
  $\vf_t $\!\!\!\!\!&$= \sigma (\mW_{if} \vx_t + \mW_{hf} \vh_{t-1})$\\
   $\vo_t $\!\!\!\!\!&$= \sigma (\mW_{io} \vx_t + \mW_{ho} \vh_{t-1})$\\
   $\vc^m_t $\!\!\!\!\!&$= \tanh (\mW_{ic} \vx_t +  \mW_{hc} \vh_{t-1})$\\
   $\vc_t $\!\!\!\!\!&$= \!  \vf_t \odot \vc_{t-1} \!+\!   \vi_t \odot \vc^m_t$\\
   $\vh_t $\!\!\!\!\!&$=\! \vo_t \odot \tanh(\vc_t)$
\end{tabular}
}} & \multicolumn{5}{l}{\multirow{9}{*}{
\begin{tabular}[l]{@{}rl@{}}
        \( A(x_t) $\!\!\!\!\!&$= \begin{bmatrix}
        \mB_t & \mD_t \\
        \mE_t & \mF_t 
       \end{bmatrix} \),\  \  \  \( g(x_t) = \begin{bmatrix}
        g_c(x_t)   \\
        g_h(x_t)   
       \end{bmatrix} \) \\
        $\mB_t$\!\!\!\!\!&$= \diag[g_f(x_t)]$\\
       $\mE_t$\!\!\!\!\!&$=\diag\left[ g_o(x_t)\odot \tanh'[g_c(x_t)]\right] \mB_t$ \\
       $\mD_t$\!\!\!\!\!&$= \diag[g^m_c(x_t) \odot  f_i(x_t)] \mW_{hi}$  \\
       \textcolor{white}{$\mD_t$}\!\!\!\!\!&$+ \diag[g_i(x_t) \odot  f^m_c(t)]  \mW_{hc}$ \\
       $\mF_t$\!\!\!\!\!&$=\diag\left[ g_o(x_t)\odot \tanh'[g_c(x_t)]\right]  \mD_t$ \\
       \textcolor{white}{$\mD_t$}\!\!\!\!\!&$+\diag\left[ f_o(x_t) \odot \tanh[g_c(x_t)] \right] \mW_{ho}$\\
\end{tabular}
}} & \multicolumn{5}{l}{\multirow{9}{*}{
\begin{tabular}[l]{@{}rl@{}}
       where:\\
       $g_c(x_t)$\!\!\!\!\!&$=g_i(x_t) \odot g^m_c(x_t)$,\\
       $g_h(x_t)$\!\!\!\!\!&$= g_o(x_t)\odot \tanh[g_c(x_t)]$,\\
      $g_i(x_t)$\!\!\!\!\!&$ = \sigma(\mW_{ii} \vx_t), \ \ \ \ \ \ \ \ \ f_i(x_t)=g'_i(x_t)$,\\
      $g_f(x_t)$\!\!\!\!\!&$=\sigma(\mW_{if} \vx_t), \ \ \ \ \ \ \ \ \!f_f(x_t)=g'_f(x_t)$,\\
      $g_o(x_t)$\!\!\!\!\!&$=\sigma(\mW_{io} \vx_t), \ \ \ \ \ \ \ \ f_o(x_t)=g'_o(x_t)$,\\
      $g^m_c(x_t)$\!\!\!\!\!&$=\tanh(\mW_{ic} \vx_t)$,\\
      $f^m_c(x_t)$\!\!\!\!\!&$=\tanh'(\mW_{ic} \vx_t)$.
\end{tabular}
}} \\
                      & \multicolumn{4}{l}{}                  & \multicolumn{5}{l}{}                  & \multicolumn{5}{l}{}                  \\
                      & \multicolumn{4}{l}{}                  & \multicolumn{5}{l}{}                  & \multicolumn{5}{l}{}                  \\
                      & \multicolumn{4}{l}{}                  & \multicolumn{5}{l}{}                  & \multicolumn{5}{l}{}     \\
                      & \multicolumn{4}{l}{}                  & \multicolumn{5}{l}{}                  & \multicolumn{5}{l}{}                  \\
                      & \multicolumn{4}{l}{}                  & \multicolumn{5}{l}{}                  & \multicolumn{5}{l}{}                  \\
                      & \multicolumn{4}{l}{}                  & \multicolumn{5}{l}{}                  & \multicolumn{5}{l}{}                  \\
                      & \multicolumn{4}{l}{}                  & \multicolumn{5}{l}{}                  & \multicolumn{5}{l}{}                  \\
                      & \multicolumn{4}{l}{}                  & \multicolumn{5}{l}{}                  & \multicolumn{5}{l}{}                  \\
                      \bottomrule
\end{tabular}
}
\vspace{-3mm}
\caption{Parameterization of $A$ and $g$ by Elman RNN, GRU, and LSTM. $\vx_t$ is the representation of the input token $x_t$ and $\mW_{**}$ refers to a weight matrix. $\sigma$ and $\tanh$ are the \textit{element-wise} {sigmoid} and $\tanh$ functions respectively. $g'$, $\tanh'$ and $f'$ refer to the \textit{element-wise} derivative. The $\diag$ operation converts a vector into a diagonal matrix.}
\label{tab:rnn_definitions}
\vspace{-4mm}
\end{table*}

\subsection{Model Parameterization}

With the above understandings on such salient components within RNNs, we can now look into how different variants of RNNs parameterize the functions $A$ and $g$.
The definition of Elman RNN, GRU and LSTM together with the corresponding Jacobian matrix $A(x_t)$ and vector function $g(x_t)$ functions are listed in Table \ref{tab:rnn_definitions}\footnote{For brevity, we suppress biases following \citet{merrill-etal-2020-formal2}.}.
We discuss how such different parameterizations may lead to different expressive power when they are used in practice.


We can see the ways GRU or LSTM parameterize  $A(x_t)$ and $g(x_t)$ appear to be more complex compared to Elman RNN.
This can partially be attributed to their gating mechanisms.
Although the original main motivation of introducing such mechanisms may be to alleviate the exploding gradient and vanishing gradient issues {\color{black}\citep{hochreiter1997long, cho-etal-2014-learning}}, 
we could see such designs also result in terms describing gates and intermediate representations.
$A$ and $g$ are then independently derived based on certain rich interactions between such terms.
We believe such interactions may likely increase the expressive power of the resulting $n$-gram representations.
We will validate these points and discuss more in our experiments.

\section{Experiments}

In our experiments, we focus on the following aspects: 1) understanding the effectiveness of the proposed $n$-gram (and context) representations when used in practice, as compared to baseline models; 2) examining the significance of the choice of context representation; 3) interpreting the proposed representations by examining how well they could be used to capture certain linguistic phenomena.

\textcolor{black}{
We employ the sentiment analysis, relation classification, named entity recognition (NER) and language modeling tasks as testbeds. The first task is often used in investigating $n$-gram phenomena {\color{black}\citep{yessenalina-cardie-2011-compositional, li-etal-2016-visualizing}} while the others are often used in examining how capable an encoder is when extracting features from texts {\color{black}\citep{grave2016improving,zhou-etal-2016-attention,lample-etal-2016-neural}}. 
}

\paragraph{Datasets}
For sentiment analysis, we considered the Stanford Sentiment Treebank (SST) \citep{socher2013recursive}, the IMDB \citep{maas-etal-2011-learning2}, and the AG-news topic classification\footnote{
AG-news can be viewed as a special sentiment analysis dataset.} \citep{NIPS2015_250cf8b5} datasets. The first dataset has sufficient labels for phrase-level analysis, the second dataset has instances with relatively longer lengths, and the third one is multi-class.
\textcolor{black}{For relation classification and NER, we considered the SemEval 2010 Task 8 \citep{hendrickx-etal-2010-semeval} and CoNLL-2003 \citep{tjong-kim-sang-de-meulder-2003-introduction}} datasets respectively.
For language modeling, we considered the Penn Treebank (PTB) dataset \citep{marcus-etal-1993-building}, the Wikitext-2 (Wiki2) dataset and the Wikitext-103 (Wiki103) dataset \citep{merity2016pointer}. PTB is relatively small while Wiki103 is large.
The statistics are shown in Tables \ref{tab:sentiment_analysis_data} and \ref{tab:language_modeling_datasets} in the appendix.

\paragraph{Baselines}
{
%
The $n$-gram representations {\color{black}(together with their corresponding context representations)} discussed in the literature are considered as baselines, which are listed in Table \ref{tab:compositional_functions_comparison} along with the MVMA and MVM models.
\textcolor{black}{MVM(A)-G/L/E refers to the MVM(A) model created with the $A$ and $g$ functions derived from GRU/LSTM/Elman, but are trained directly from data. The $A$ and $g$ functions for GRU, LSTM and Elman are listed in Table \ref{tab:rnn_definitions}.}
}

Additionally, \textcolor{black}{to understand whether the complexity of $A$ affects the expressive power}, we created a new model called MVMA-ME, which comes with an $A$ function that is slightly more complex than that of MVMA-E but less complex than those of MVMA-G and MVMA-L:
$A(x_t)\!=\!0.25\diag[\tanh(\mW \vx_t)] \mM\!+\!0.5\mI$ and $g(x_t)=\tanh(\mW' \vx_t)$ (here, $\mW$, $\mM$ and $\mW'$ are learnable weight matrices). The $g$ function is the same as that of MVMA-E.



\paragraph{Setup}

\textcolor{black}{For sentiment analysis, relation classification and language modeling, models consist of one embedding layer, one RNN layer, and one fully-connected layer. 
The {Adagrad} optimizer \citep{duchi2011adaptive} was used along with  {dropout} \citep{srivastava2014dropout} for sentiment analysis\footnote{We investigated the approximation between RNNs and their corresponding recurrence relations in Appendix \ref{subsect:approx_rnn_error}. The spectral normalization \citep{miyato2018spectral} was used on the weight matrices $\mW_{h*}$ for standard RNNs.} and relation classification.}
\textcolor{black}{For language modeling, models were trained with the {Adam} optimizer \citep{Adam}.
We ran word-level models with {\color{black}truncated backpropagation through time \citep{6797135truncatedBPTT} where the truncated length was set to 35.} 
{Adaptive softmax} \citep{joulin2017efficient} was used for Wiki103.}
\textcolor{black}{For NER, models consist of one embedding layer, one bidirectional RNN layer, one projection layer and one conditional random field (CRF) layer. The SGD optimizer was used.}
Final models were chosen based on the best validation results. More implementation details can be found in the appendix.

\begin{table}[t!]
\centering
\scalebox{0.5}{
\begin{tabular}{lcccccc}
\toprule
\multicolumn{1}{c}{\multirow{2}{*}{\textbf{Model}}} & \multicolumn{2}{c}{\textbf{SST-2}}                   & \multicolumn{2}{c}{\textbf{AG-news}}                 & \multicolumn{2}{c}{\textbf{IMDB}}                    \\
\multicolumn{1}{c}{}                                & \multicolumn{1}{c}{dev} & \multicolumn{1}{c}{test} & \multicolumn{1}{c}{dev} & \multicolumn{1}{c}{test} & \multicolumn{1}{c}{dev} & \multicolumn{1}{c}{test} \\ 
\midrule
MM                                                  & 86.0$\pm$1.3              & 85.6$\pm$0.4             & -                         & -                        & \multicolumn{1}{c}{-}     & \multicolumn{1}{c}{-}    \\
VA-W                                                & 80.6$\pm$1.6              & 80.4$\pm$1.4             & 90.3$\pm$0.4              & 90.0$\pm$0.3             & 88.0$\pm$0.6              & 88.0$\pm$0.4             \\
VA-EW                                               & 82.6$\pm$0.3              & 82.0$\pm$0.3             & -                         & -                        & \multicolumn{1}{c}{-}     & \multicolumn{1}{c}{-}    \\
MVM-G                                               & 84.9$\pm$0.5              & 85.0$\pm$1.0             & 84.9$\pm$4.0              & 84.4$\pm$4.0             & 50.9$\pm$0.0              & 50.2$\pm$0.1             \\
MVM-L                                               & 85.4$\pm$0.4              & 84.9$\pm$0.8             & 86.9$\pm$1.7              & 86.5$\pm$1.7             & 51.0$\pm$0.1              & 50.2$\pm$0.1             \\
MVM-E                                               & 59.6$\pm$1.6              & 59.5$\pm$1.1             & -                         & -                        & \multicolumn{1}{c}{-}     & \multicolumn{1}{c}{-}    \\
\midrule
MVMA-G                                              & 87.0$\pm$0.4              & 85.3$\pm$0.5             & 91.6$\pm$0.5              & 91.3$\pm$0.3             & 90.5$\pm$0.5              & 89.6$\pm$0.7             \\
MVMA-L                                              & 86.7$\pm$1.0              & 85.4$\pm$1.0             & 91.4$\pm$0.5              & 91.3$\pm$0.5             & 89.4$\pm$0.6              & 89.2$\pm$0.6             \\
MVMA-E                                              & 81.4$\pm$1.1              & 80.8$\pm$1.5             & -                         & -                        & \multicolumn{1}{c}{-}     & \multicolumn{1}{c}{-}    \\
MVMA-ME                                           & 83.2$\pm$0.5              & 81.9$\pm$0.3             & 90.6$\pm$0.5              & 90.2$\pm$0.3             & 80.6$\pm$0.5              & 80.1$\pm$1.1             \\
\midrule
GRU                                                 & 84.9$\pm$0.9              & 84.9$\pm$0.5             & 92.1$\pm$0.1              & 91.6$\pm$0.3             & 87.7$\pm$0.2              & 87.2$\pm$0.3             \\
LSTM                                                & 84.3$\pm$0.8              & 84.4$\pm$0.3             & 91.9$\pm$0.4              & 91.5$\pm$0.5             & 89.0$\pm$0.1              & 88.7$\pm$0.4             \\
Elman                                               & 79.1$\pm$0.3              & 79.7$\pm$1.4             & 87.5$\pm$0.5              & 87.5$\pm$0.6             & 67.0$\pm$1.9              & 66.7$\pm$0.9             \\
\bottomrule
\end{tabular}
}
\vspace{-2mm}
\caption{Accuracy percentage ($\uparrow$) on sentiment analysis (text classification) datasets (averaged over 3 runs). ``-'' means the model failed to converge.}
\vspace{-2mm}
\label{tab:accuracy_compositional_models}
\vspace{-2mm}
\end{table}

\subsection{Comparison of Representation Models}
\label{sec:component_expressivity}
We investigate how baseline $n$-gram representation models\footnote{We excluded VM, which we found was hard to train. We also excluded MVM-R which only considers bigrams.}, \textcolor{black}{the MVM model, and the MVMA model perform on the aforementioned testbeds.}
We also compare with the standard RNN models.

\paragraph{Sentiment Analysis}
Apart from the GRU and LSTM models, it can be observed that our MVMA-G and MVMA-L models are also able to achieve competitive results on three sentiment analysis datasets, as we can see from Table \ref{tab:accuracy_compositional_models}, demonstrating the efficacy of those recurrence-induced $n$-gram representations.
Although Elman RNN and its corresponding {\color{black}MVMA-E and MVM-E models} also have a mechanism for capturing $n$-gram information (similar to GRU and LSTM), they did not perform well, which may be attributed to a limited expressive power of their $A$ and $g$ functions when used for defining $n$-grams as described previously.

{Both   MM  and  VA-EW  fail to converge on AG-news and IMDB, showing challenges for them to handle long instances. This may be explained by the lengthy matrix multiplication involved in their representations, which may result in vanishing/exploding gradient issues.} 
Interestingly,  MVM-G and MVM-L, which solely rely on the longest $n$-gram representation, are also able to achieve good results on SST-2, indicating a reasonable expressive power of such $n$-gram representations alone. However, they fail to catch up with  MVMA-G and MVMA-L  on IMDB which contains much longer instances, confirming the significance of the context representation, which captures $n$-grams of varying lengths.

Unlike MVMA-E, the MVMA-ME model does not suffer from loss stagnation on  AG-news and IMDB but the performance on IMDB obviously falls behind MVMA-G and MVMA-L as shown in Table \ref{tab:accuracy_compositional_models}. This indicates a sufficiently expressive $A(x_t)$ (such as the Jacobian matrices of GRU and LSTM) may be needed to handle long instances.

\paragraph{Relation Classification \& NER}
\textcolor{black}{
For relation classification, context representations (or  final hidden states) are used for classification. For NER, we use the concatenated context representations (or hidden states) at each position of bidirectional models to predict entities and their types.
Table \ref{tab:f1_scores_rc_ner} shows that  MVMA-G and MVMA-L  outperform the MVM-G and MVM-L models respectively on both tasks, again confirming the effectiveness of the context representations. MVM(A)-E did not perform as well as MVM(A)-G and MVM(A)-L, which demonstrates the significance of expressive power for the $A$ and $g$ functions.
Similar to the results in sentiment analysis, MVMA-ME did not perform as well as MVMA-G and MVMA-L. However, to our surprise, MVMA-ME did not outperform VA-EW on NER, suggesting that a delicate choice of $A$ can be important for this task. 
The poor performance of VA-W on NER might be explained by a weak expressive power of its $n$-gram representations. MM fails to converge on the relation classification task, which implies it is not robust across different datasets.
Interestingly, it is remarkable that MVMA-G, MVMA-L and MVMA-E could yield competitive results compared to GRU, LSTM and Elman on NER, implying such $n$-gram representations could be crucial for our NER task. }

\begin{table}[]
\centering
\scalebox{0.5}{
\begin{tabular}{lllll}
\toprule
\multicolumn{1}{c}{\multirow{2}{*}{\textbf{Model}}} & \multicolumn{2}{c}{\textbf{Relation Classification}} & \multicolumn{2}{c}{\textbf{NER}} \\
\multicolumn{1}{c}{}                                & \multicolumn{1}{c}{dev}  & \multicolumn{1}{c}{test}                     &  \multicolumn{1}{c}{dev}  & \multicolumn{1}{c}{test}           \\
\midrule
MM                                                  & \multicolumn{1}{c}{-}     & \multicolumn{1}{c}{-}    & 33.9$\pm$0.6    & 30.8$\pm$0.4   \\
VA-W                                                & 41.2$\pm$0.2              & 37.9$\pm$0.9             & 17.6$\pm$0.6	&16.5$\pm$1.6  \\
VA-EW                                               & 39.7$\pm$1.1              & 38.3$\pm$0.7             & 70.8$\pm$0.7    & 63.4$\pm$1.0     \\
MVM-G                                               & 51.2$\pm$0.5              & 52.6$\pm$0.7             & 54.2$\pm$1.6    & 47.6$\pm$2.2   \\
MVM-L                                               & 48.8$\pm$1.3              & 50.5$\pm$1.5             & 53.8$\pm$1.7    & 46.6$\pm$1.6   \\
MVM-E                                               & \multicolumn{1}{c}{-}     & \multicolumn{1}{c}{-}    & 27.8$\pm$0.9    & 25.6$\pm$0.9   \\
\midrule
MVMA-G                                              & 62.2$\pm$1.0              & 59.7$\pm$0.1             & 75.0$\pm$0.4      & 67.7$\pm$0.5   \\
MVMA-L                                              & 57.5$\pm$0.3              & 56.2$\pm$0.8             & 75.6$\pm$0.2    & 67.9$\pm$0.3   \\
MVMA-E                                              & 27.8$\pm$0.9              & 25.6$\pm$0.9             & 69.0$\pm$0.4    & 61.7$\pm$0.1   \\
MVMA-ME                                             & 46.3$\pm$0.9              & 46.2$\pm$0.6             & 67.0$\pm$0.5    & 57.6$\pm$0.8   \\
\midrule
GRU                                                 & 67.2$\pm$0.6              & 62.2$\pm$0.2             & 75.6$\pm$0.5    & 67.9$\pm$0.5   \\
LSTM                                                & 65.2$\pm$0.9              & 61.3$\pm$1.4             & 76.3$\pm$0.5    & 68.1$\pm$0.5   \\
Elman                                               & 27.8$\pm$0.9              & 25.6$\pm$0.9             & 67.1$\pm$0.9    & 58.6$\pm$0.6  
\\
\bottomrule
\end{tabular}
}
\vspace{-2mm}
\caption{F1 scores ($\uparrow$) (averaged over 3 runs) on the relation classification and NER tasks. ``-'' means the model failed to converge.}
\vspace{-2mm}
\label{tab:f1_scores_rc_ner}
\end{table}

\paragraph{Language Modeling}
For the language modeling task, we choose MVMA-G, MVMA-L, MVM-G and MVM-L for experiments. We also run MVMA-ME. 
As we can see from Table \ref{tab:encoder_performance}, there are performance gaps between the MVMA models and the standard RNNs -- though the gaps often do not appear to be particularly large. 
This indicates there may be extra information  within higher order terms of the standard RNN functions useful for such a task.
Yet, such information cannot be captured by the  MVMA models that employ simplified functions.
The gaps between the MVM models and MVMA models are remarkable, which again indicates that the correct way of defining the left context representation can be crucial for the task of next word prediction.
{\color{black}
 MVMA-ME  did not perform well on the language modeling task, which might be attributed to the less expressive power of its functions $A$ and $g$.
}


\begin{table}[t!]
\centering
\scalebox{0.5}{
\begin{tabular}{lrccc}
\toprule
\multicolumn{1}{c}{\multirow{1}{*}{\textbf{Model}}} && {\bf PTB}             & {\bf Wiki2}          & {\bf Wiki103}       \\
\midrule
 \multirow{2}{*}{\small GRU}  & dev  & 118.4$\pm$0.4 & 146.1$\pm$0.4 & 109.4$\pm$0.6 \\
                                         & test   & 110.1$\pm$0.4   & 136.8$\pm$0.1 & 113.3$\pm$0.8 \\\cmidrule{2-5}
                     \multirow{2}{*}{\small MVMA-G}  & dev  & 119.8$\pm$0.4  & 150.3$\pm$0.8 & 111.8$\pm$0.5 \\
                     
                                         & test   & 111.1$\pm$0.2  & 140.2$\pm$1.0 & 115.2$\pm$0.5 \\
                    \cmidrule{2-5}
                      \multirow{2}{*}{\small MVM-G}  & dev  &   146.5$\pm$1.3    &   170.1$\pm$2.8  &   -  \\
                                          & test   & 138.8 $\pm$1.0   &  160.0$\pm$2.6   &   -   \\
\midrule
 \multirow{2}{*}{\small LSTM}  & valid  & 118.6$\pm$0.4  & 150.6$\pm$0.6 & 108.3$\pm$0.6 \\
                                         & test   & 109.8$\pm$0.4  & 140.4$\pm$0.8 & 112.4$\pm$0.8 \\
                                         \cmidrule{2-5}
                     \multirow{2}{*}{\small MVMA-L}  & dev  & 121.5$\pm$0.5  & 152.0$\pm$0.5 & 109.1$\pm$0.6 \\
                                          & test   & 113.2$\pm$0.5  & 142.5$\pm$0.7 & 112.6$\pm$0.6
                    \\ \cmidrule{2-5}
                    
                       \multirow{2}{*}{\small MVM-L}  & dev  &  124.3$\pm$1.5    &  155.6$\pm$0.9  &   -  \\
                                          & test   & 117.0$\pm$1.0    & 145.7$\pm$1.6  &  - 
                    \\
\midrule
\multirow{2}{*}{\small MVMA-ME} & dev & 140.7$\pm$0.9   & 169.0$\pm$1.0 & 153.1$\pm$4.2
\\    & test  & 134.0$\pm$1.0  & 158.4$\pm$1.4 & 157.4$\pm$4.3 \\
                    
                    \bottomrule
\end{tabular}
}
\vspace{-2mm}
\caption{Perplexities ($\downarrow$) on language modeling (averaged over 5 runs). ``-'': the model failed to converge.}
\label{tab:encoder_performance}
\vspace{-2mm}
\end{table}

\subsection{Interpretation Analysis}
We conduct some further analysis to examine the interpretability of $n$-gram representations. 
Specifically, we examine whether the models are able to capture certain linguistic phenomena such as negation, which is important for sentiment analysis \cite{ribeiro-etal-2020-beyond}. We also additionally made comparisons with the vanilla Transformer \citep{NIPS2017_vaswani} here\footnote{The mean of output representations was treated as the context representation for Transformer during training. We also tried to use the concatenation of the first and last token, following \cite{luan2019general}, which yielded similar results.} despite the fact that it remains largely unclear how it precisely captures sequence features such as $n$-grams.

\textcolor{black}{We could also obtain the $n$-gram representations and the corresponding context representations from the learned standard RNN models, based on their learned parameters. We denote such $n$-gram representations as RNN\textsubscript{$n$-gram}, and the context representations as RNN\textsubscript{context}, where ``RNN'' can be GRU, LSTM or Elman.}
\textcolor{black}{As $n$-gram representations are vectors, a common approach is to transform them into scalars with learnable  parameters \citep{DBLP:journals/corr/abs-1801-05453murdoch,sun-lu-2020-understanding2}. 
We define the \textit{$n$-gram polarity score} to quantify the polarity information as captured by an $n$-gram representation $\vv_{i:t}$ from time step $i$ to $t$, which is calculated as:
\begin{equation}
\small
    \begin{aligned}
       s^{\vv}_{i:t} = \vw^\top \vv_{i:t},
    \end{aligned}
\end{equation}
where $\vw$ is the learnable weight vector of the final fully-connected layer. 
We also define the \textit{context polarity score} for the {context} as $\sum_{i=1}^{t}s^{\vv}_{i:t}$.}

\begin{figure*}[t!]
\centering
\vspace{-4mm}
    \begin{subfigure}{0.23\textwidth}
      \centering
      \includegraphics[scale=0.2]{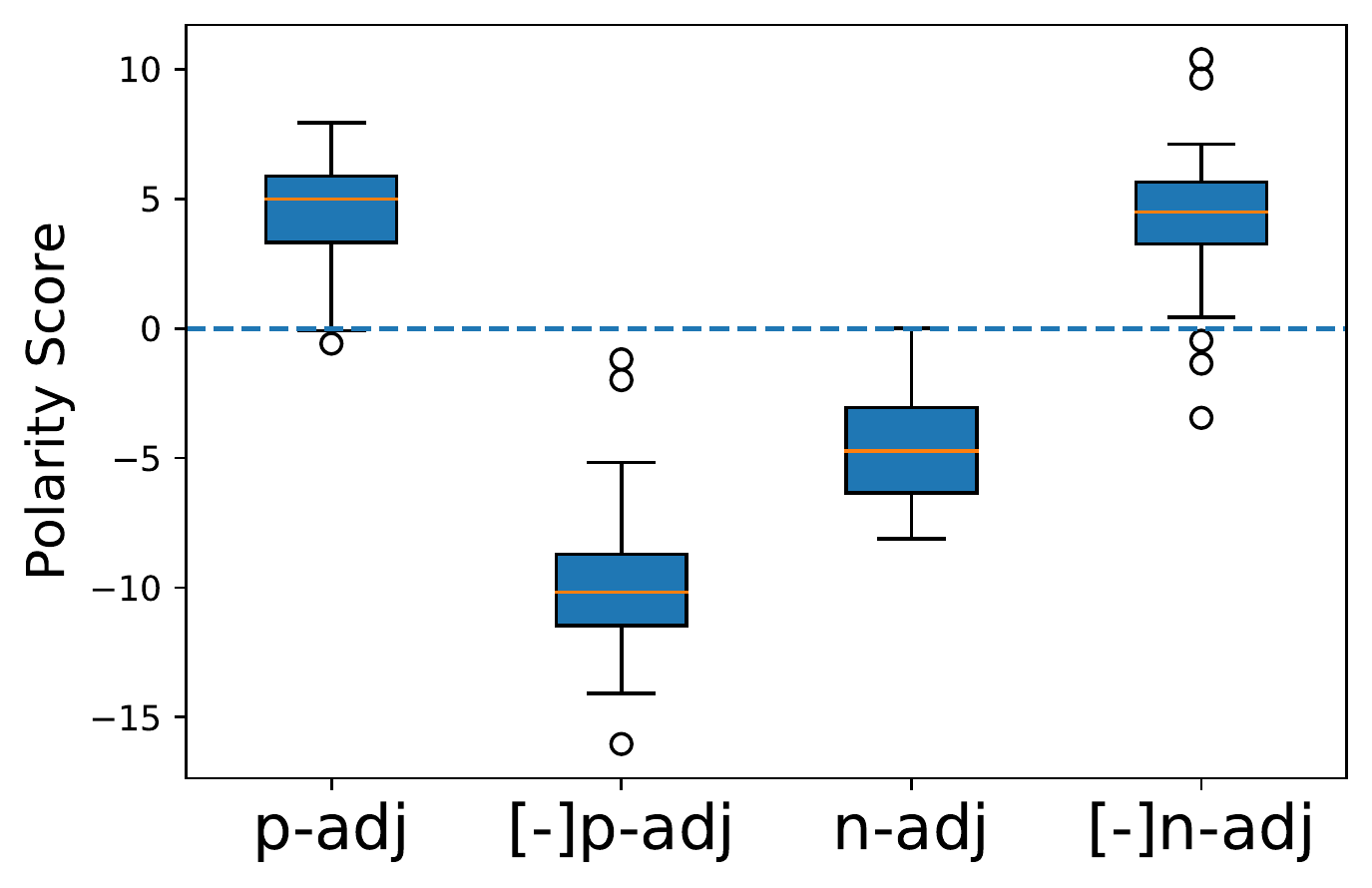}
      \vspace{-2mm}
      \caption{GRU\textsubscript{$n$-gram}}
      \label{fig:gru_standard}
    \end{subfigure}
    \begin{subfigure}{0.23\textwidth}
      \centering
      \includegraphics[scale=0.2]{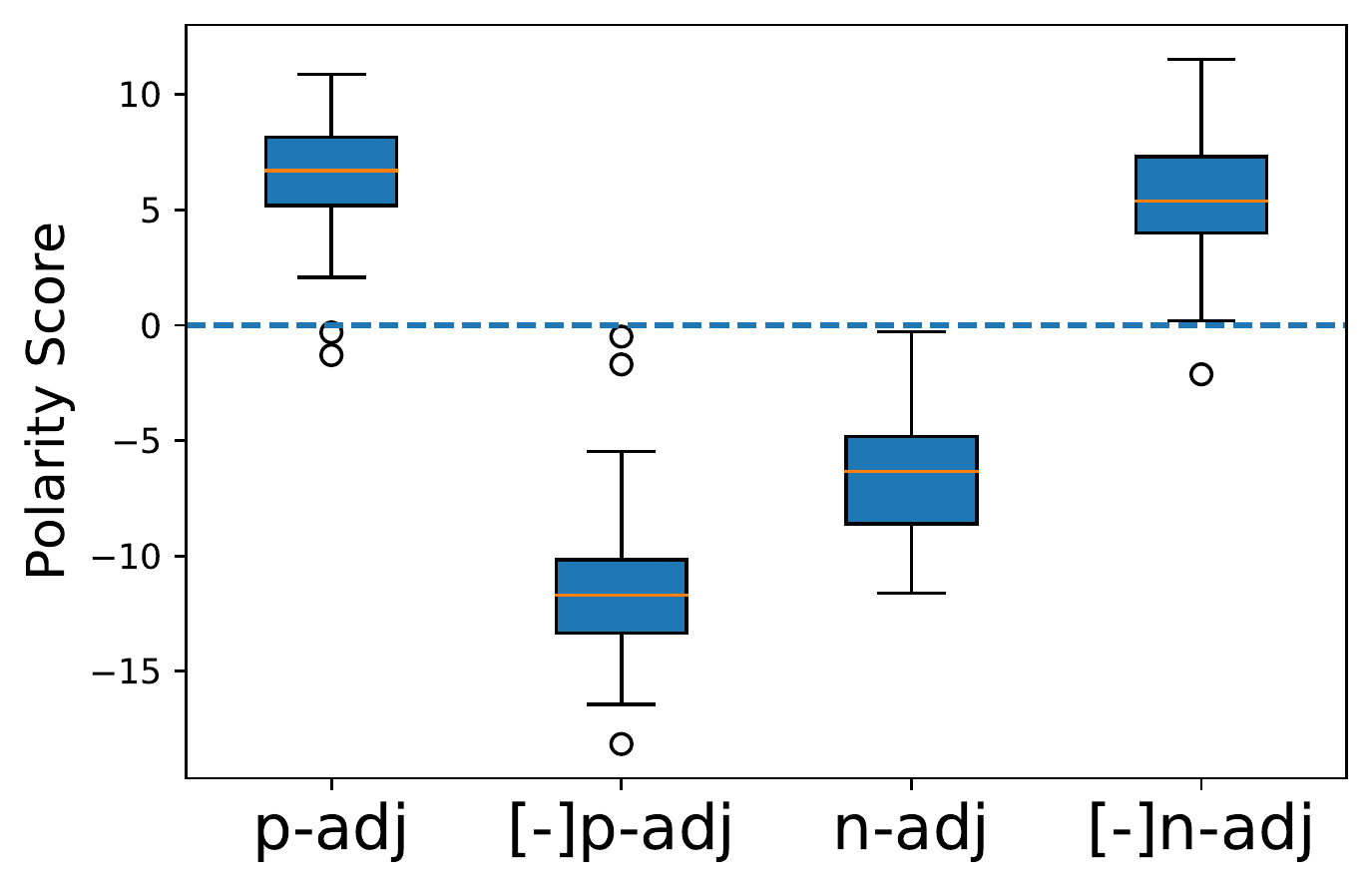}
      \vspace{-2mm}
      \caption{MVMA-G}
      \label{fig:gru_o_n_gram}
    \end{subfigure}
    \begin{subfigure}{0.23\textwidth}
      \centering
      \includegraphics[scale=0.2]{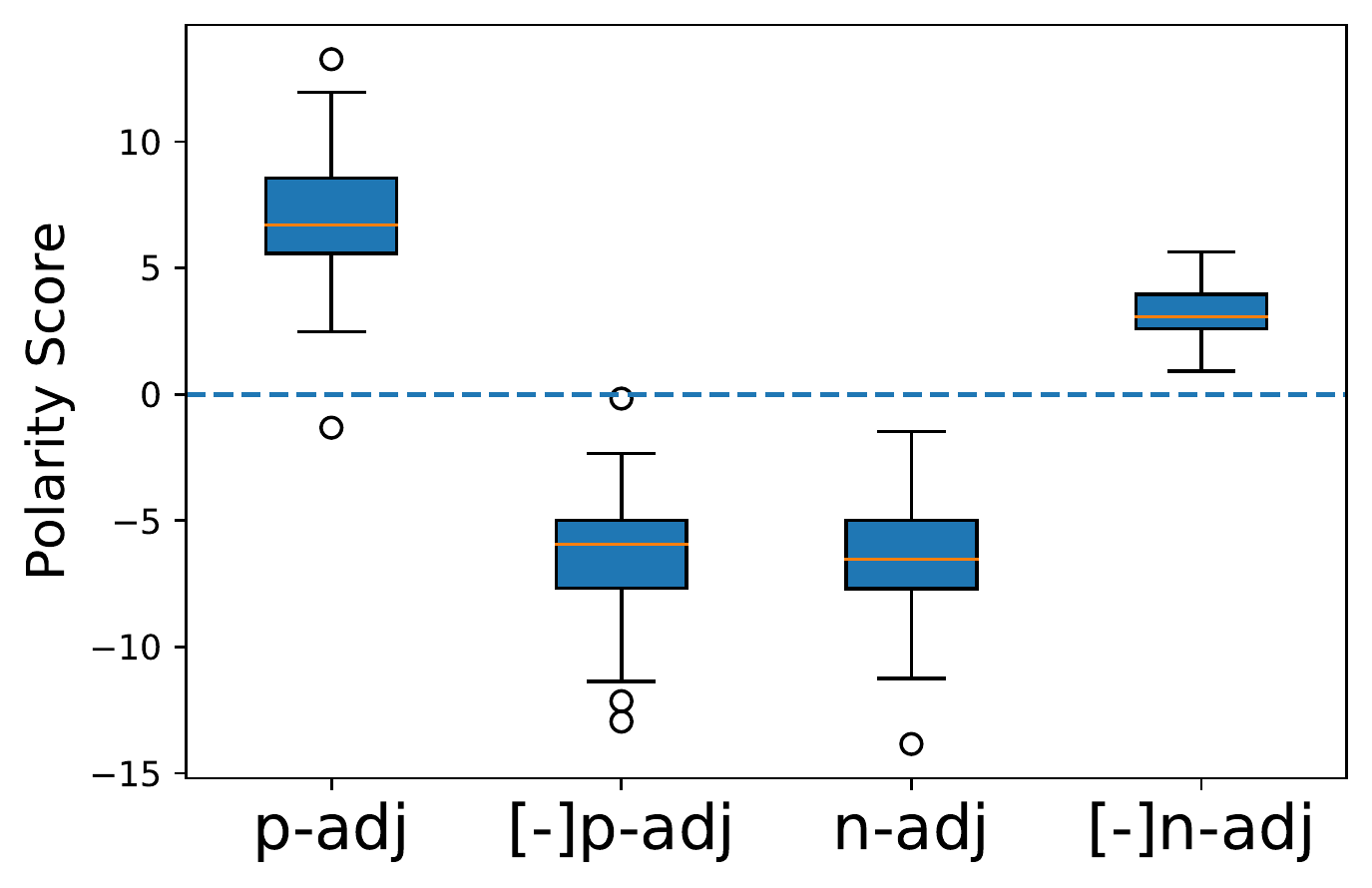}
      \vspace{-2mm}
      \caption{MVM-G}
      \label{fig:gru_l_n_gram}
    \end{subfigure}
    \begin{subfigure}{0.23\textwidth}
      \centering
      \includegraphics[scale=0.2]{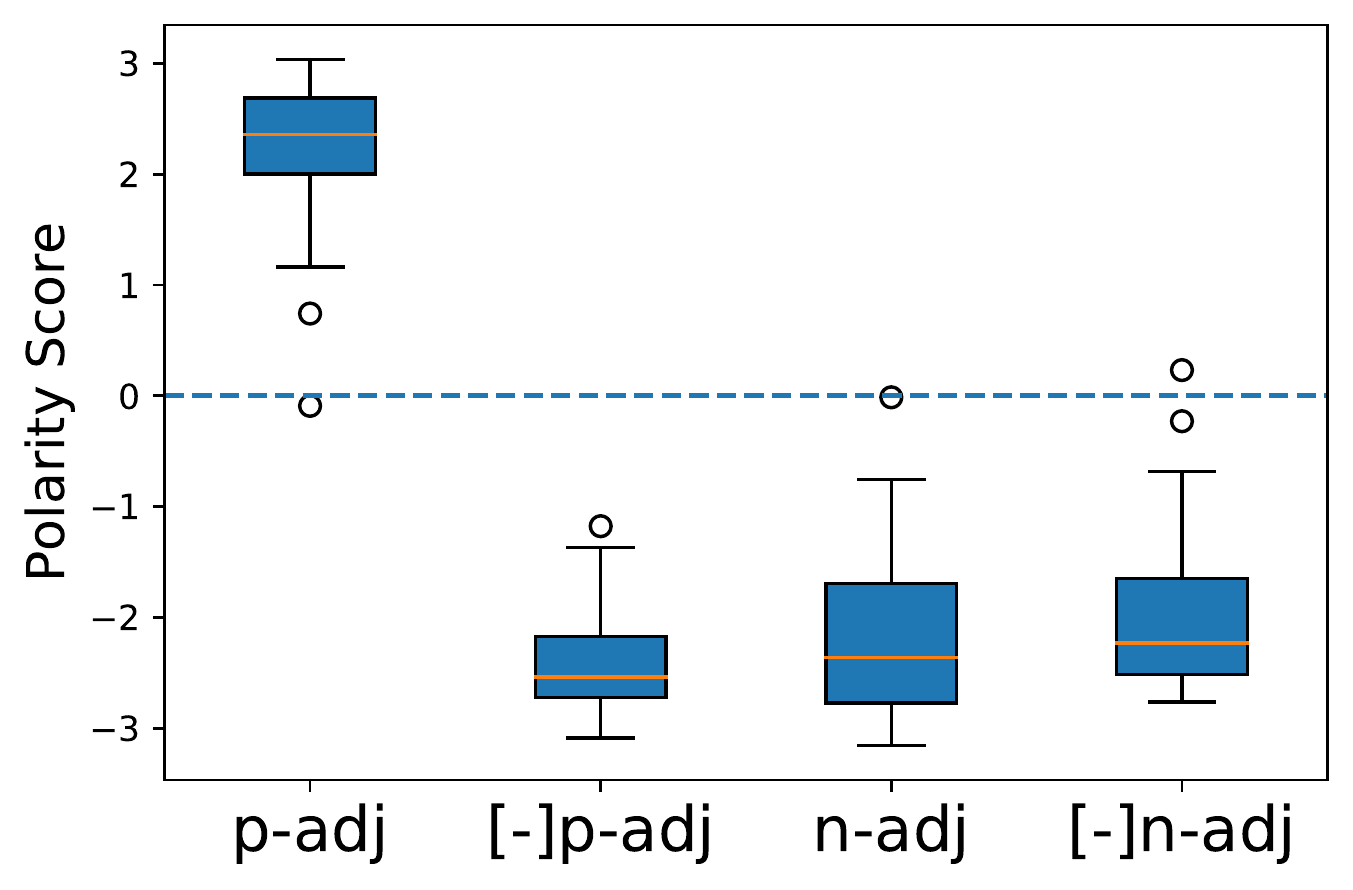}
      \vspace{-2mm}
      \caption{MVMA-E}
      \label{fig:elman_o_n_gram}
    \end{subfigure}
    \begin{subfigure}{0.23\textwidth}
      \centering
      \includegraphics[scale=0.2]{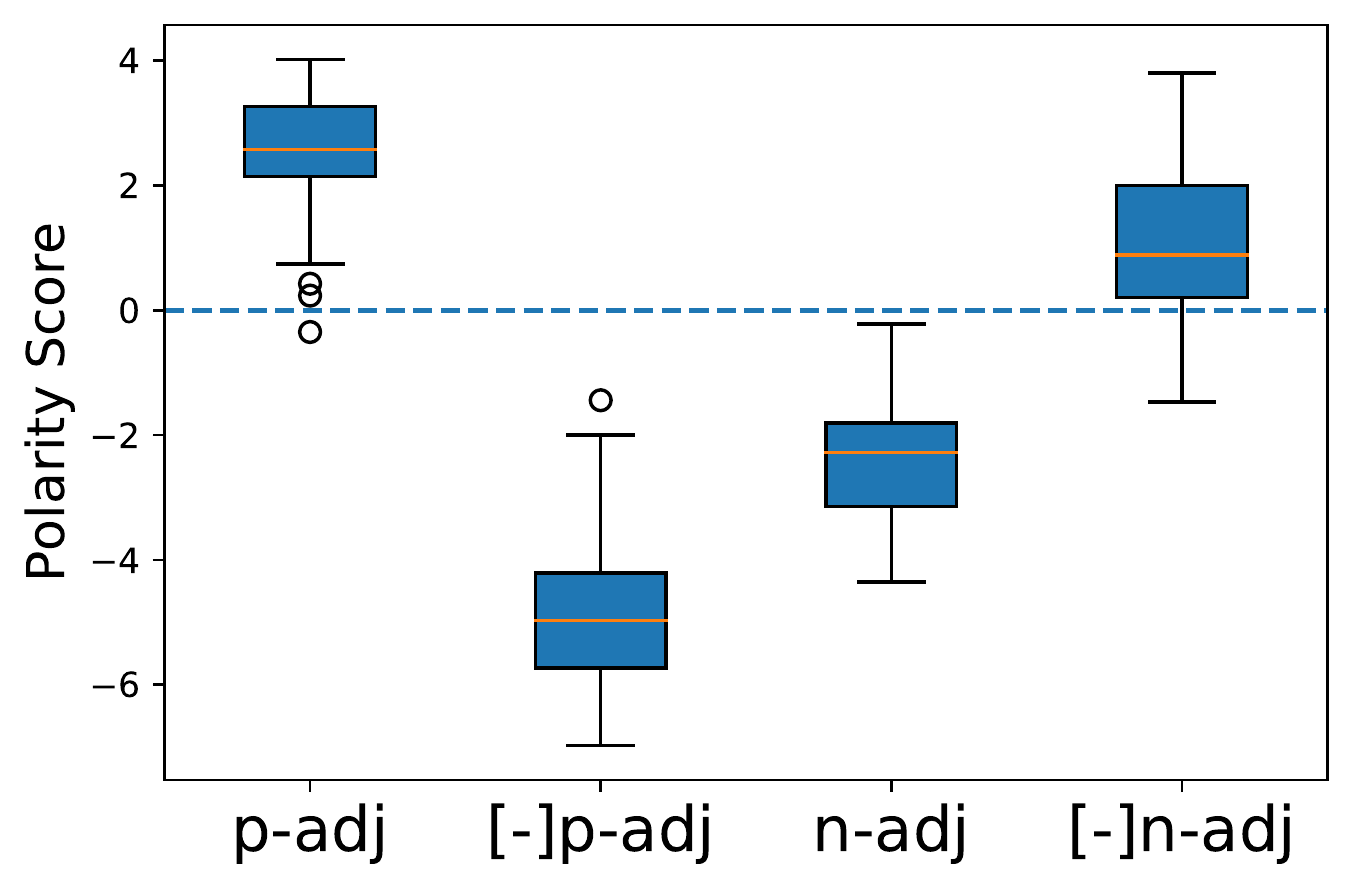}
      \vspace{-2mm}
      \caption{MVMA-ME}
      \label{fig:linear_ii}
    \end{subfigure}
    \begin{subfigure}{0.23\textwidth}
      \centering
      \includegraphics[scale=0.2]{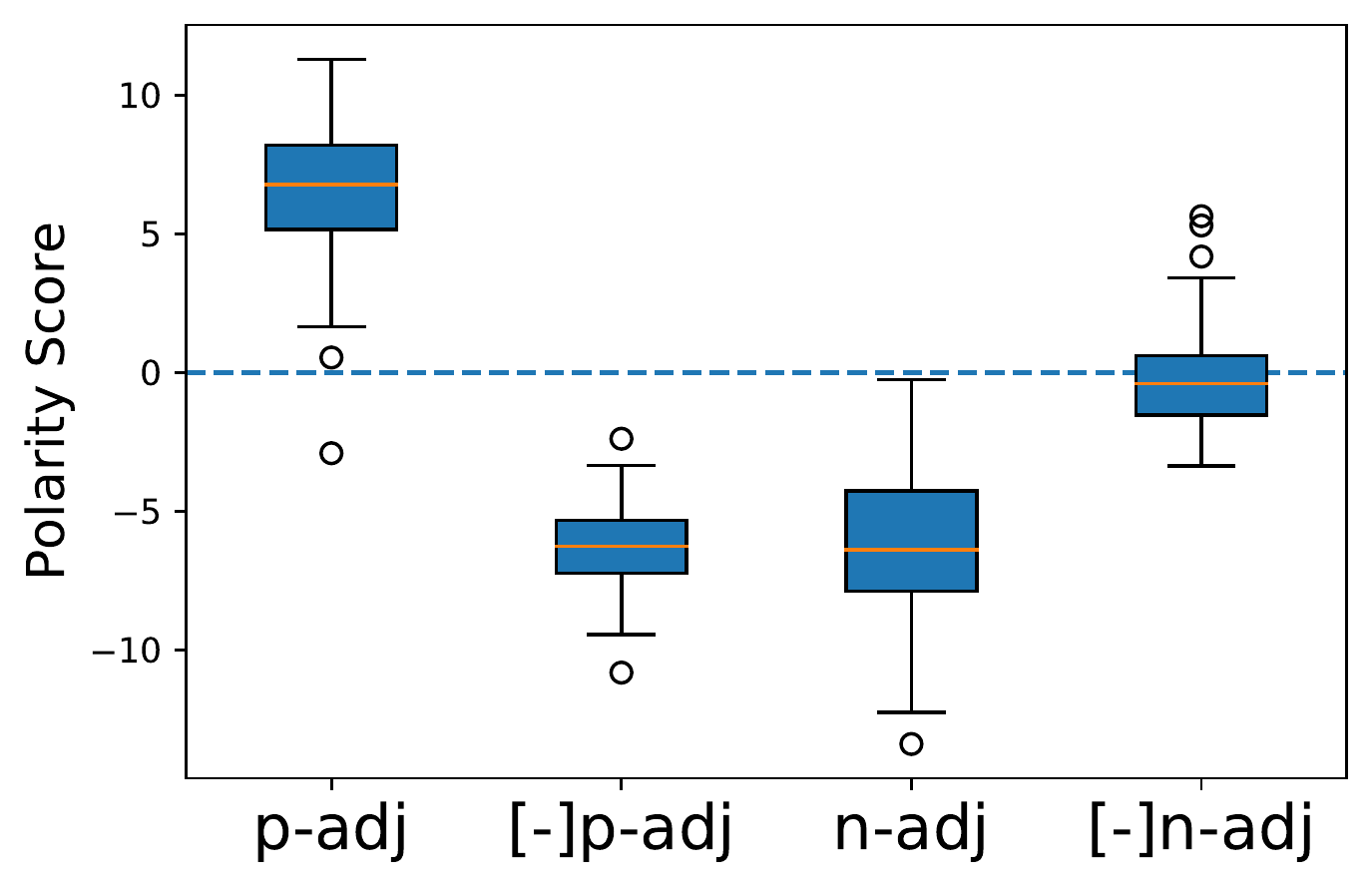}
      \vspace{-2mm}
      \caption{MM}
      \label{fig:maxtrix_space}
    \end{subfigure}
    \begin{subfigure}{0.23\textwidth}
      \centering
      \includegraphics[scale=0.2]{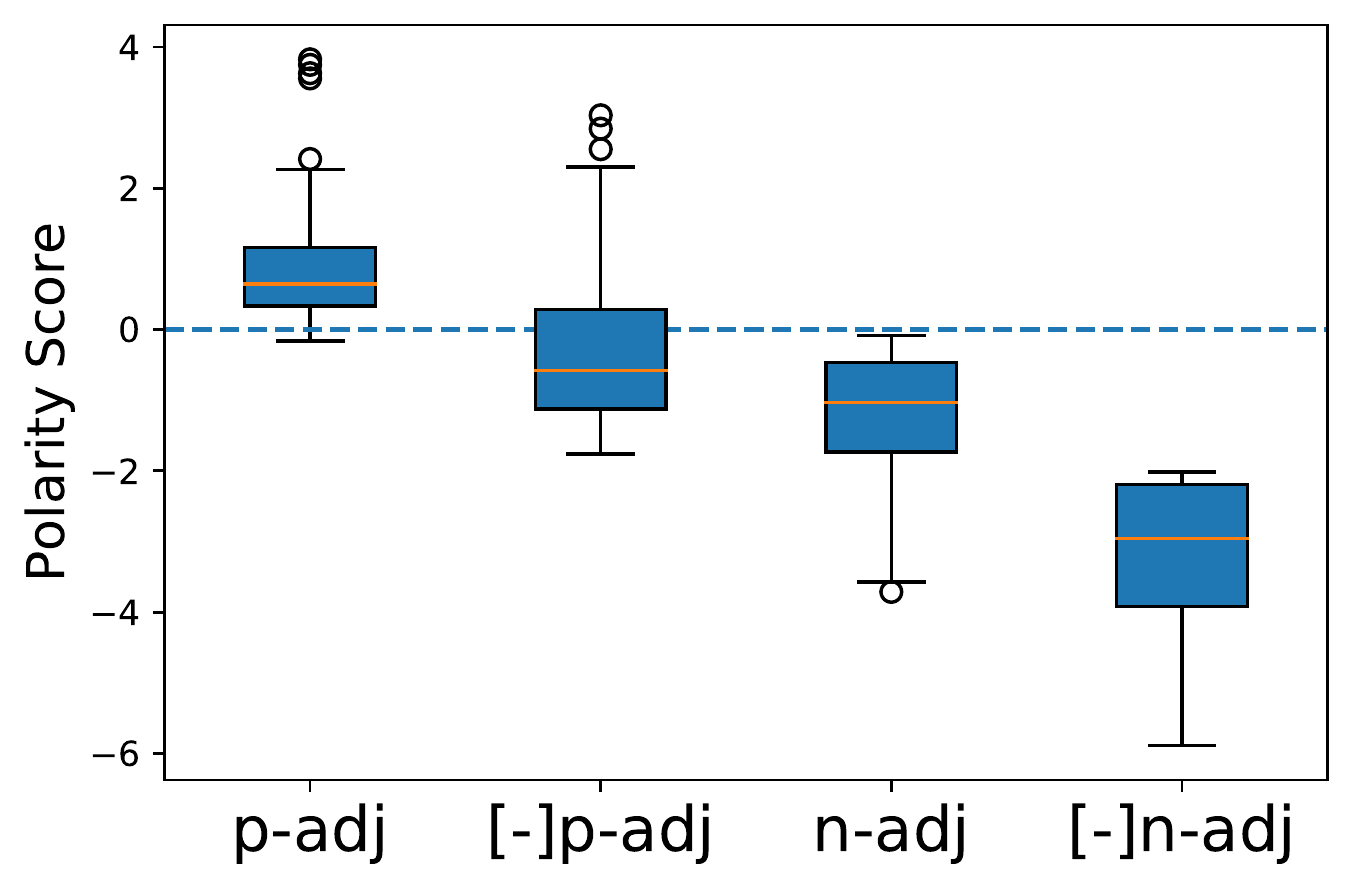}
      \vspace{-2mm}
      \caption{VA-W}
      \label{fig:additive_model}
    \end{subfigure}
    \begin{subfigure}{0.23\textwidth}
      \centering
      \includegraphics[scale=0.2]{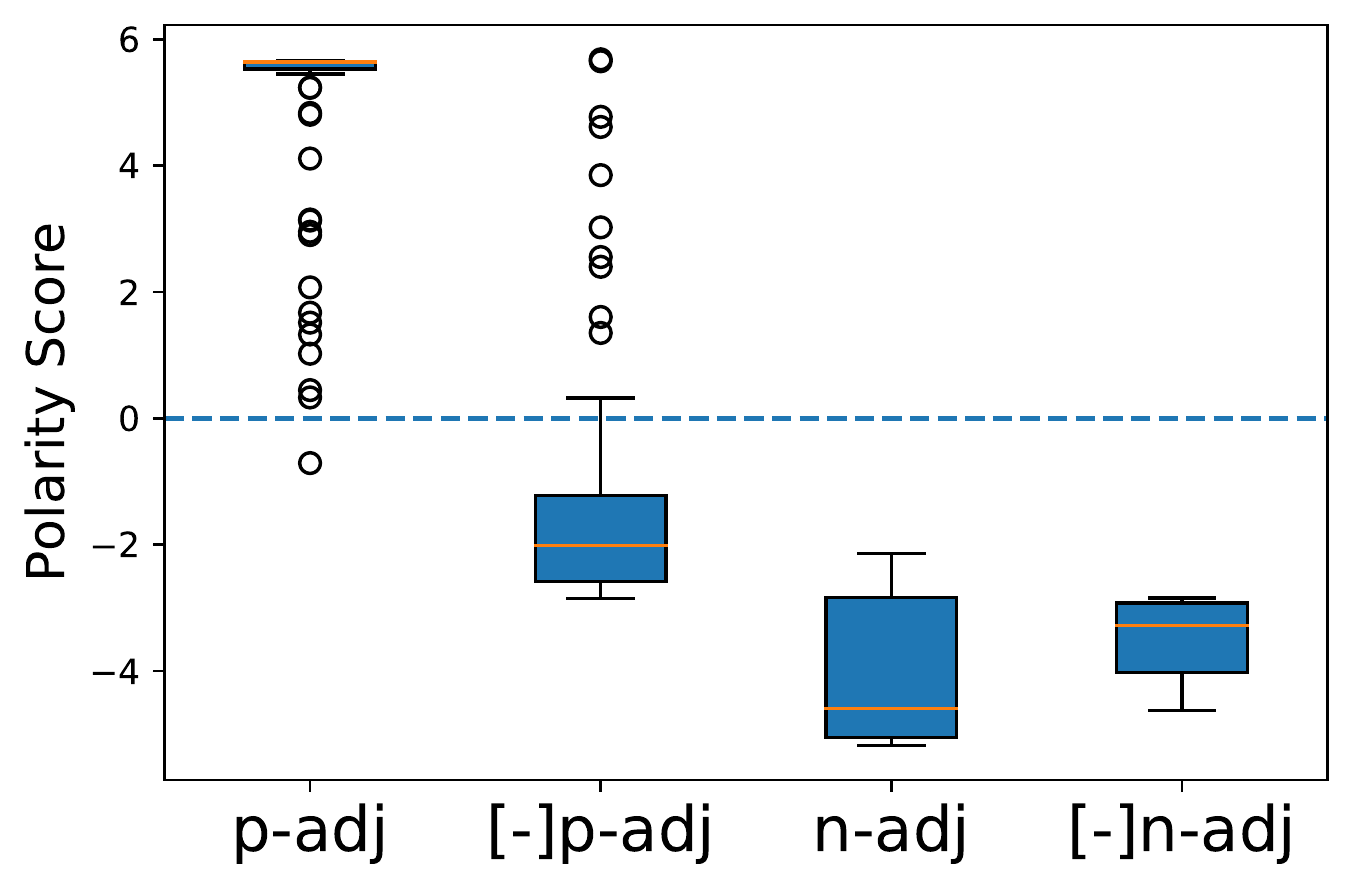}
      \vspace{-2mm}
      \caption{Transformer}
      \label{fig:transformer}
    \end{subfigure}
    \vspace{-2mm}
    \caption{Distribution of polarity scores for adjectives and their negation bigrams on SST-2. \textit{p-adj} and \textit{n-adj} refer to the positive and negative adjectives respectively.  [-] refers to the negation operation (prepending the word ``not''). Circles refer to outliers. More results can be found in the appendix.}
    \label{fig:sst_rnn_negate_adj}
    \vspace{-3mm}
\end{figure*}

{\color{black}We trained  RNNs and baseline models on  SST-2 and automatically extracted 73 positive adjectives (e.g., ``{\em nice}'' and ``{\em enjoyable}'') and 47 negative adjectives (e.g., ``{\em bad}'' and ``{\em tedious}'') from the vocabulary\footnote{Such adjectives and detailed automatic extraction process can be found in the appendix.}. $N$-gram polarity scores were calculated for those adjective unigrams and their negation bigrams formed by prepending  ``{\em not}'' to them. For  VA-EW and VA-W, their $n$-gram representations do not  involve tokens other than the last token.
Such limitations prevent them from capturing any negation information.
We therefore calculate the  context polarity scores using their context representations instead (which in this case is a bigram). This also applies to Transformer for the same reason.}

We observed that, for the GRU and LSTM models along with their corresponding {MVMA} models, the $n$-gram representations are generally able to learn the negation for both the adjective and their negation bigrams as shown in Figures \ref{fig:gru_standard} and \ref{fig:gru_o_n_gram}\footnote{Results of  LSTM  are similar to GRU, which can be found in the appendix.}, prepending ``{\em not}'' to an adjective will likely reverse the polarity. 
This might be a reason why they could achieve relatively higher accuracy on the sentiment analysis tasks.
Interestingly,  MVM-G could also capture  negation as shown in Figure \ref{fig:gru_l_n_gram}, again suggesting the impressive expressive power of such $n$-gram representations alone.

However, as shown in Figure \ref{fig:sst_rnn_negate_adj}, models such as  VA-W, MVMA-E, and MM  \textcolor{black}{are struggling to capture  negation for negative adjectives, again implying a weaker expressive power of their $n$-gram representations}.
Specifically, MVMA-E fails to capture  negation for negative adjectives, \textcolor{black}{which may be attributed to a relatively weaker Jacobian matrix function $A$ (as compared to those of GRU and LSTM)  preventing them from pursuing optimal conditions. }

{\color{black}
Figure \ref{fig:linear_ii} shows that the MVMA-ME model, which has a function $A$ less complex than the ones from MVMA-G and MVMA-L but more complex than the one from MVMA-E, still can generally learn  negation of negative adjectives better than the MVMA-E model.
This demonstrates the necessity of choosing more expressive $A$ and $g$ functions.

}

\textcolor{black}{Interestingly, both  VA-W and Transformer are struggling with capturing the negation phenomenon for negative adjectives} in our experiments as shown in Figures \ref{fig:additive_model} and \ref{fig:transformer}, which suggests that they may have a weaker capability in modeling sequential features in our setup. However, we found they could still achieve good performances on the AG-news and IMDB datasets\footnote{\color{black}We conducted additional experiments for Transformers on sentiment analysis. Results are in appendix.}. 
We hypothesize this is because the  nature of  SST-2  makes these two models suffer more on this dataset -- it has rich linguistic phenomena such as negation cases while the other two datasets do not.



\begin{figure}[t!]
\centering
    \vspace{-2mm}
\scalebox{1.12}{
\hspace{-2mm}
      \includegraphics[scale=0.25]{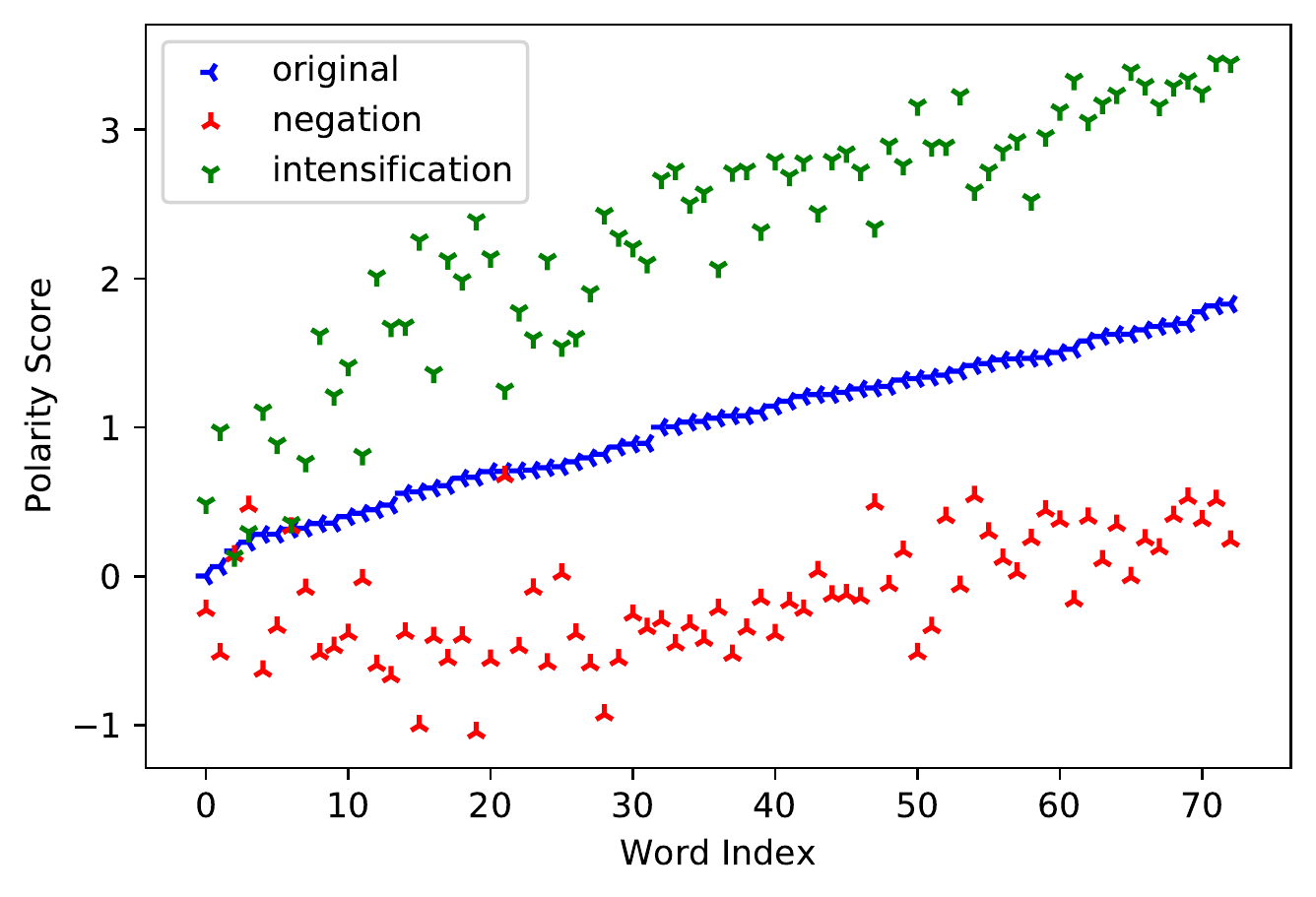}
      \includegraphics[scale=0.25]{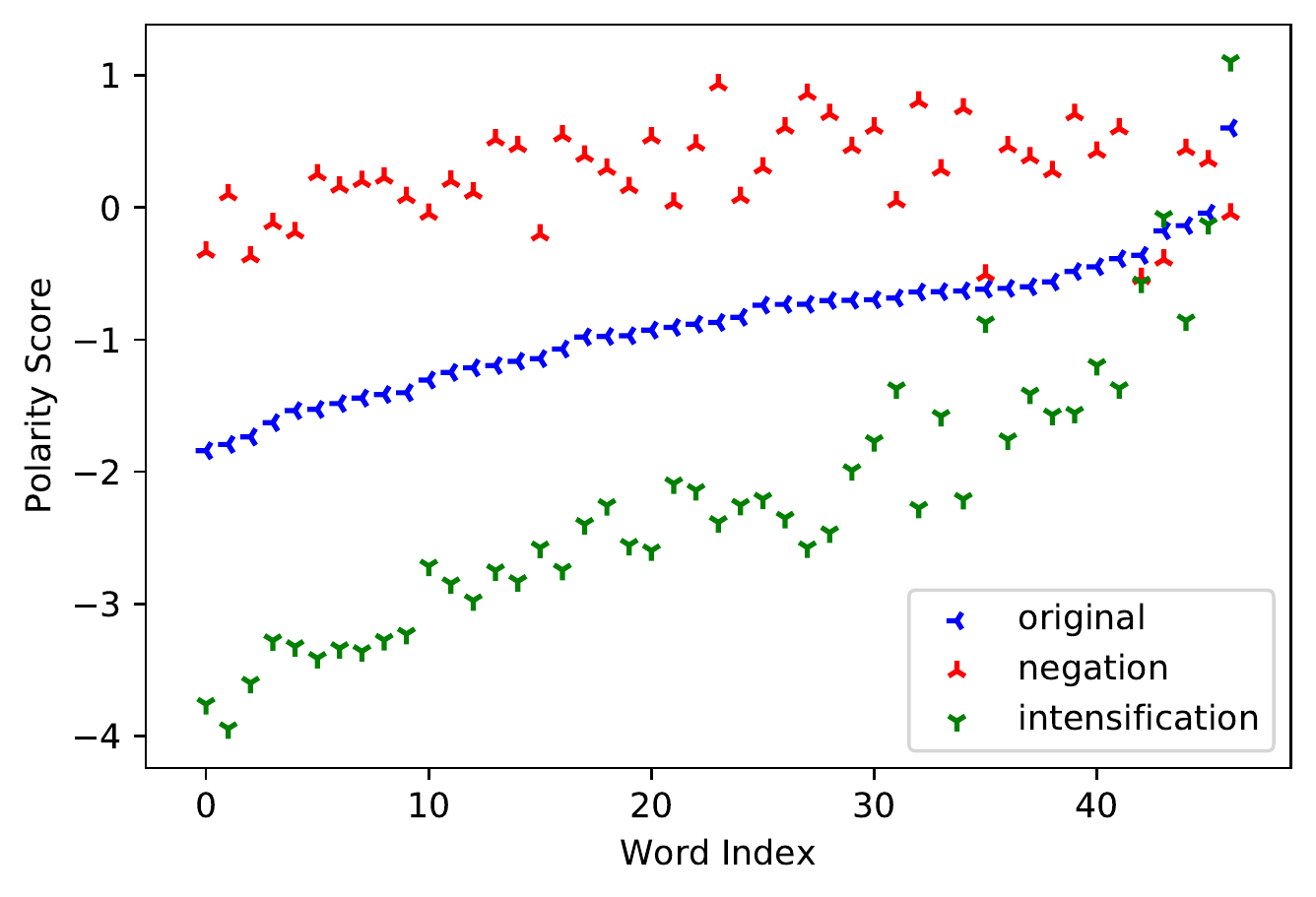}}
    \vspace{-2mm}
    \vspace{-2mm}
    \vspace{-2mm}
    \vspace{-1mm}
          \caption{\color{black}Context polarity scores (MVMA-G, SST-5) for positive  (L) and negative  (R) adjectives along with their  negation and intensification bigrams.}
    \label{fig:gru_negation_intensification_example}
    \vspace{-2mm}
    \vspace{-2mm}
\end{figure}


{\color{black}
Additionally, we examined the ability for GRU, LSTM, MVMA-G and MVMA-L to capture both the negation and intensification phenomena.
For such experiments, instead of using SST-2, we trained the models on SST-5, which comes with polarity intensity information.
Polarity intensities were mapped into values of $\{-2, -1, 0, +1, +2\}$, ranging from \textit{extremely negative} to  \textit{extremely positive}.
We conducted some experiments based on the same setup above for capturing negation on SST-2. To our surprise, our preliminary results show that all models were  performing substantially worse in terms of capturing intensification than capturing negations.
We hypothesize that this is caused by the imbalance between negation phrases and intensification phrases. Specifically, the intensification word {\em``very''} (1,729 times) was exposed less than the negation word {\em ``not''} (4,601 times) in the training set of SST-5.
}

{\color{black}One approach proposed in the literature for sentence classification  is to consider all the hidden states of an RNN in an instance \citep{Bahdanau2015NeuralMT}.
We believe this may actually be able to alleviate the above issue as it allows  more $n$-grams within an instance to be exposed to the label information. Thus, we followed their approach for training our MVMA and MVM models\footnote{However, for simplicity, in this work we only used the mean context representations (or hidden states) instead of a weighted sum of them.}.}

We can see that the negation and intensification phenomena can be explained by both the context representations in Figure \ref{fig:gru_negation_intensification_example}\footnote{More results are in the appendix. }.
Specifically, {\color{black}prepending either positive or negative adjectives with ``{\em very}''} will likely strengthen their polarity while adding ``{\em not}'' will likely weaken their polarity.
These results suggest that RNNs are able to capture information of linguistic significance  within the sequence, and our identified $n$-gram representations within their hidden states appear to be playing a salient role.



\subsection{Discussion}

From the experiments above, we can see that our introduced $n$-gram representations, coupled with the corresponding context representations, are powerful in practice  in capturing $n$-gram information better than the baseline compositional models introduced in the literature.
We also found that RNNs can induce such representations due to their recurrence mechanism\footnote{We also visualized the context representations and $n$-gram representations in the appendix, which provide intuitive understanding of them.}.

However, there can be several factors that affect the efficacy of different representations.
First, through comparisons with different variants of MVMA, we can see that the precise way of parameterizing the  functions $A(x_t)$ and $g(x_t)$ matter.
Second, through the comparison between MVMA and MVM, we can see that defining an appropriate context representation that incorporates a correct set of $n$-grams is also important.
Third, for models which do not capture such explicit $n$-gram features like ours, interestingly, they may still be able to yield good performances on certain tasks.
For example,  though VA-W and Transformer  did not perform well on  SST-2, they yielded  results competitive to  GRU and LSTM  on  AG-news and IMDB.
This observation indicates  there could be other useful features captured by such models that can contribute towards  their overall modeling power.

Although in this work we did not aim to propose novel or more powerful architectures, we believe our work can be a step towards better understanding of RNN models. We also hope it can provide inspiration for our community to design more interpretable yet efficient architectures.

\section{Conclusion}

In this work, we focused on investigating the underlying mechanism of RNNs in terms of handling sequential information from a theoretical perspective.
Our analysis reveals that RNNs  contain a mechanism
where each hidden state encodes a weighted combination of salient components,
each of which can be interpreted as a representation of a classical $n$-gram.
Through a series of comprehensive empirical studies on different tasks, we confirm our understandings on such interpretations of these components.
With the analysis coupled with experiments, we provide findings on how RNNs learn to handle certain linguistic phenomena such as negation and intensification.
Further investigations on understanding how the identified mechanism may capture a wider range of linguistic phenomena such as multiword expressions \citep{schneider-etal-2014-discriminative} could an interesting future direction.

\section*{Acknowledgements}

We would like to thank the anonymous reviewers
and our ARR action editor for their constructive
comments. 
This research/project is supported by the Ministry of Education, Singapore, under its Tier 3 Programme (The Award No.: MOET32020-0004).
Any opinions, findings and conclusions or recommendations expressed in this material are those of the authors and do not reflect the views of the Ministry of Education, Singapore.

\bibliography{anthology,acl}
\bibliographystyle{acl_natbib}

\newpage

\appendix
\input{appendix}

\end{document}

%% file: math_commands.tex
\usepackage{amsmath,amsfonts,bm}

\def\eqref#1{equation~\ref{#1}}

\def\1{\bm{1}}

\def\vc{{\bm{c}}}

\def\vf{{\bm{f}}}
\def\vg{{\bm{g}}}
\def\vh{{\bm{h}}}
\def\vi{{\bm{i}}}

\def\vn{{\bm{n}}}
\def\vo{{\bm{o}}}

\def\vr{{\bm{r}}}

\def\vu{{\bm{u}}}
\def\vv{{\bm{v}}}
\def\vw{{\bm{w}}}
\def\vx{{\bm{x}}}

\def\vz{{\bm{z}}}

\def\mA{{\bm{A}}}
\def\mB{{\bm{B}}}
\def\mC{{\bm{C}}}
\def\mD{{\bm{D}}}
\def\mE{{\bm{E}}}
\def\mF{{\bm{F}}}

\def\mI{{\bm{I}}}

\def\mM{{\bm{M}}}

\def\mW{{\bm{W}}}

\DeclareMathAlphabet{\mathsfit}{\encodingdefault}{\sfdefault}{m}{sl}
\SetMathAlphabet{\mathsfit}{bold}{\encodingdefault}{\sfdefault}{bx}{n}

\def\sR{{\mathbb{R}}}

\def\sV{{\mathbb{V}}}



%% file: appendix.tex
\section{Dataset Statistics}
The statistics of the sentiment analysis, relation classification and NER datasets are shown in Table \ref{tab:sentiment_analysis_data}.
The language modeling datasets are obtained from \href{https://blog.einstein.ai/the-wikitext-long-term-dependency-language-modeling-dataset}{Einstein.ai} and the statistics are shown in Table \ref{tab:language_modeling_datasets}. 
\begin{table}[!htbp]
\centering
\scalebox{0.6}{
\begin{tabular}{lrrrrrr}
        \toprule
\textbf{Data}      & \textbf{Train} & \textbf{Dev}  &  \textbf{Test} &\textbf{V.size} &\textbf{Max.len}  &\textbf{Class}          \\
\midrule
\small SST-2       &98,794  &872 &1,821    &17,404 &54 &2   \\

\small IMDB  & 17,212  &4,304  &4,363   &63,311 &437 &2  \\
\small AG-news  &   110,000   &10,000  &7,600   &85,568 & 212 &4 \\

\small SST-5 &  318,582  &41,447 &82,600  &18,025 & 54  &5   
\\
\small SemEval &  7,000  &1,000 &2,717  &27,115  & 91    & 10 
\\
\small CoNLL-2003 &  14,987  &3,466 &3,684  &26,873  &113    &  20
\\
\bottomrule
\end{tabular}
}
\vspace{-2mm}
\caption{Statistics of the sentiment analysis, relation classification and NER datasets. ``V.size'' indicates the vocabulary size and ``Max.len'' indicates the maximum length of the instances.  ``SemEval'' refers to the SemEval 2010 Task 8 dataset for relation classification. For CoNLL-2003, ``class'' refers to the tag size.}
\label{tab:sentiment_analysis_data}
\vspace{-2mm}
\end{table}


We created the binary dataset SST-2 by extracting instances (including phrases) with polarity from the constituency parse trees in the original SST dataset \citep{socher2013recursive}. We merged the labels {\em extremely positive} and {\em positive} as {\em  positive} and the labels {\em extremely negative} and {\em negative} as {\em negative}.
We also extracted all the phrases in the constituency parse trees from the original dataset and created the 5-class dataset SST-5. The labels \textit{extremely positive}, \textit{positive}, \textit{neutral}, \textit{negative} and \textit{extremely negative} were mapped into +2, +1, 0, -1, and -2 respectively.

\begin{table}[]
\scalebox{0.75}{
\begin{tabular}{cccll}
\toprule
\multicolumn{2}{c}{\textbf{Dataset}}  & \textbf{Train} & \textbf{Dev} & \textbf{Test} \\
\midrule
\multirow{2}{*}{PTB}     & Token Num  & 887,521        & 70,390       & 78,669        \\
                         & Vocab Size & \multicolumn{3}{c}{10,000}                    \\
                         \midrule
\multirow{2}{*}{Wiki2}   & Token Num  & 2,088,628      & 217,646      & 245,569       \\
                         & Vocab Size & \multicolumn{3}{c}{33,278}                    \\
                         \midrule
\multirow{2}{*}{Wiki103} & Token Num  & 103,227,021    & 217,646      & 245,569       \\
                         & Vocab Size & \multicolumn{3}{c}{267,735}  
                         \\
                         \bottomrule
\end{tabular}
}
\caption{Statistics of the language modeling datasets.}
\label{tab:language_modeling_datasets}
\end{table}

\section{More Result from the SST datasets}
\label{subsect:more_results_sst}

\subsection{Negation and Intensification}
\begin{figure}[!htbp]
\centering
\vspace{-0mm}
    \begin{subfigure}{0.23\textwidth}
      \centering
      \includegraphics[scale=0.255]{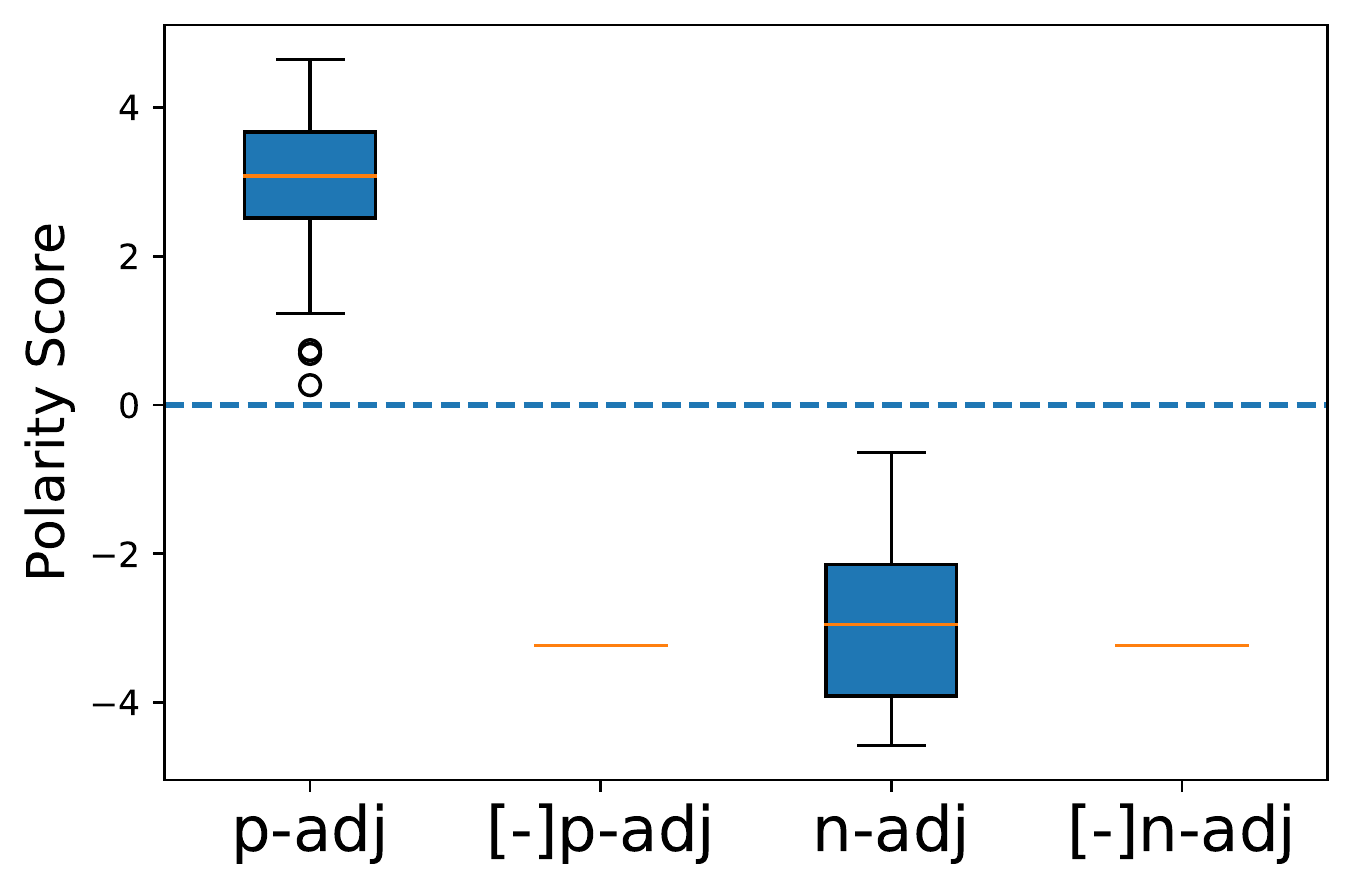}
      \vspace{-2mm}
      \caption{VA-EW}
    \end{subfigure}
    \begin{subfigure}{0.23\textwidth}
      \centering
      \includegraphics[scale=0.255]{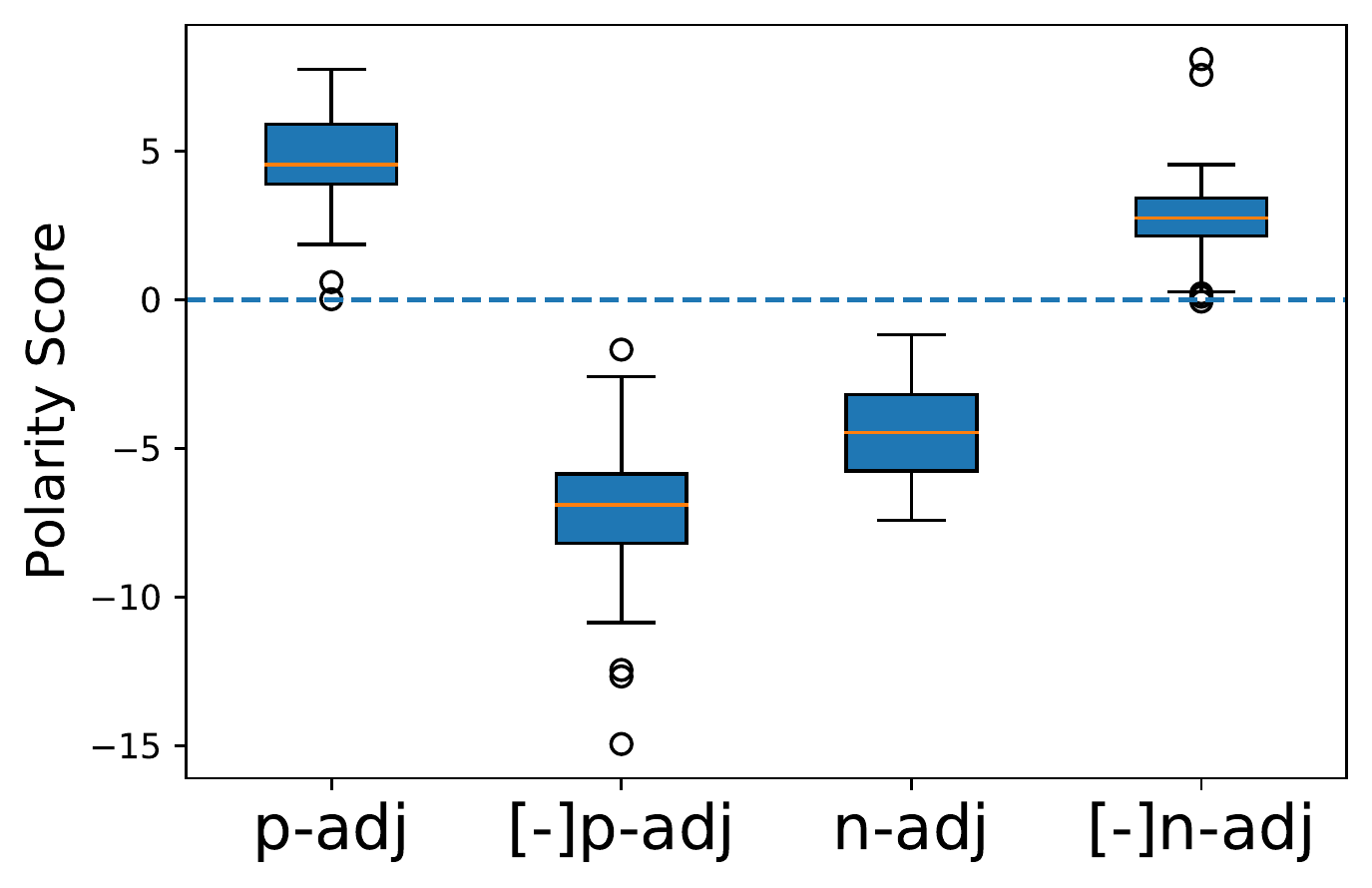}
      \vspace{-2mm}
      \caption{LSTM\textsubscript{$n$-gram}}
    \end{subfigure}
    \begin{subfigure}{0.23\textwidth}
      \centering
      \includegraphics[scale=0.255]{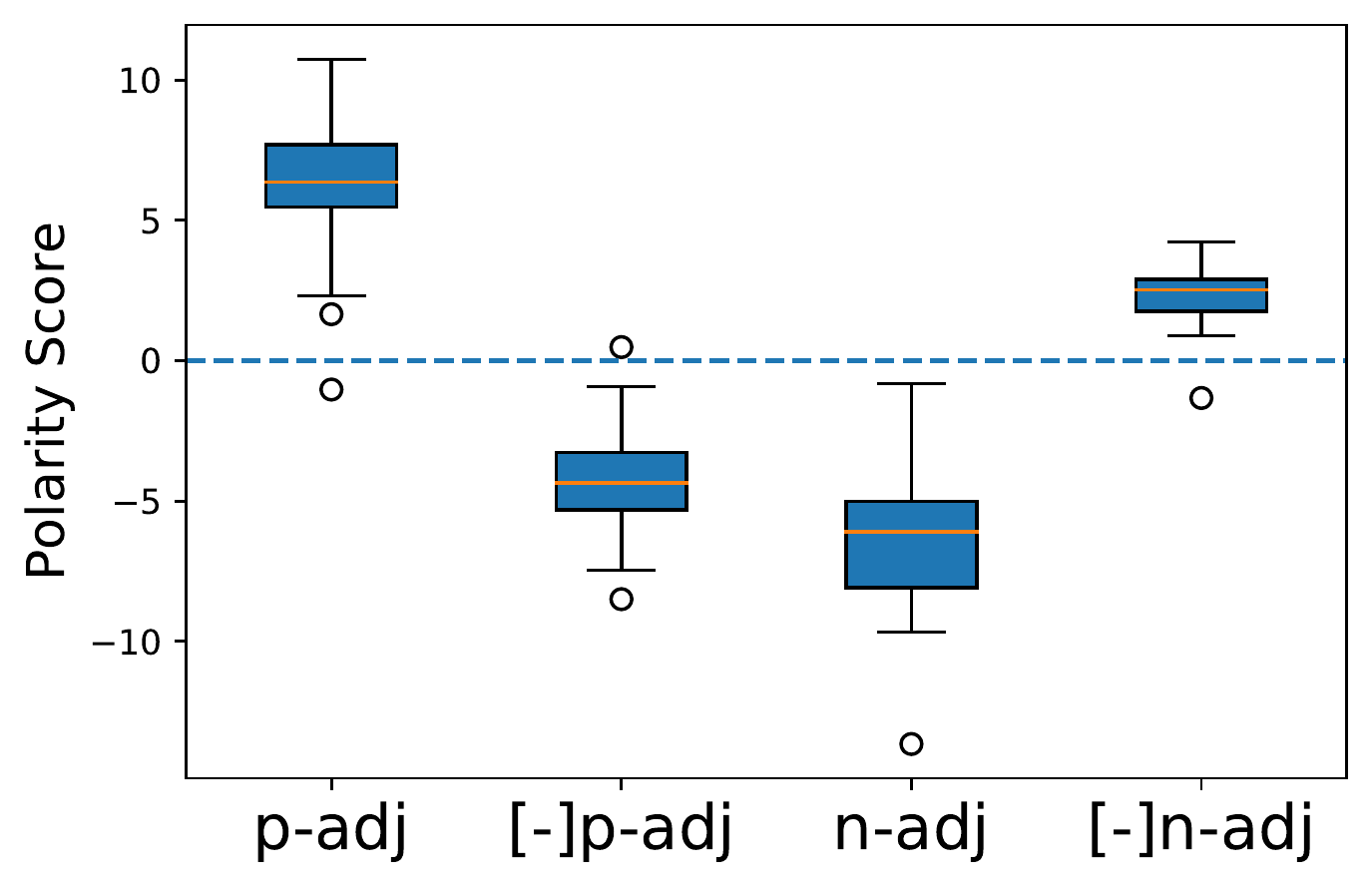}
      \vspace{-2mm}
      \caption{MVM-L}
    \end{subfigure}
    \begin{subfigure}{0.23\textwidth}
      \centering
      \includegraphics[scale=0.255]{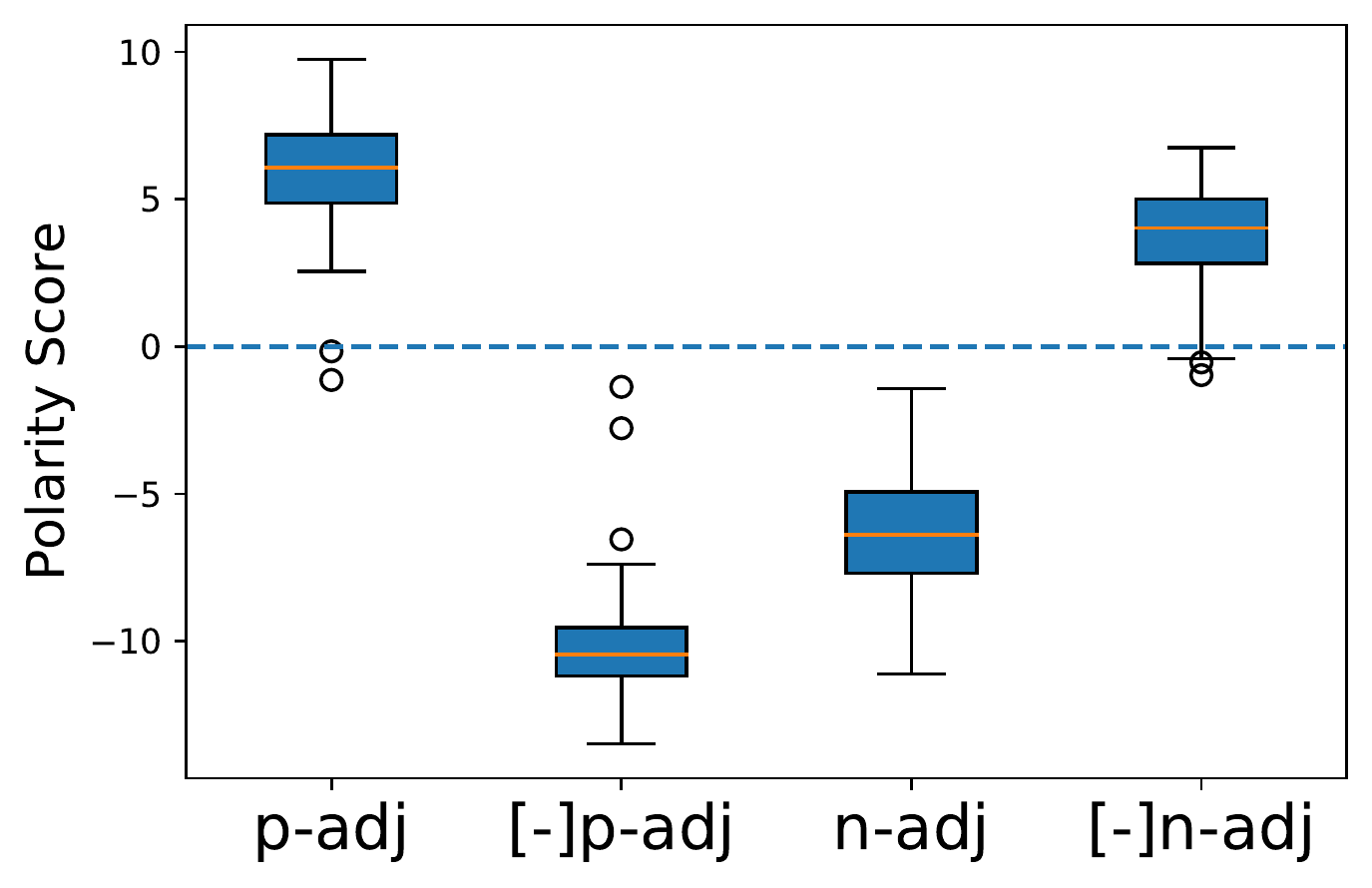}
      \vspace{-2mm}
      \caption{MVMA-L}
    \end{subfigure}
    \vspace{-2mm}
    \caption{Distribution of polarity scores for adjectives and their negation bigrams. \textit{p-adj} and \textit{n-adj} refer to the positive and negative adjectives respectively.  [-] refers to the negation operation (prepending the word ``not''). Circles refer to outliers. }
    \label{fig:lstm_negate_adjectives}
    \vspace{-3mm}
\end{figure}

\begin{figure}[t!]
\centering
\vspace{-0mm}
    
    \begin{subfigure}{0.43\textwidth}
      \centering
      \includegraphics[scale=0.4]{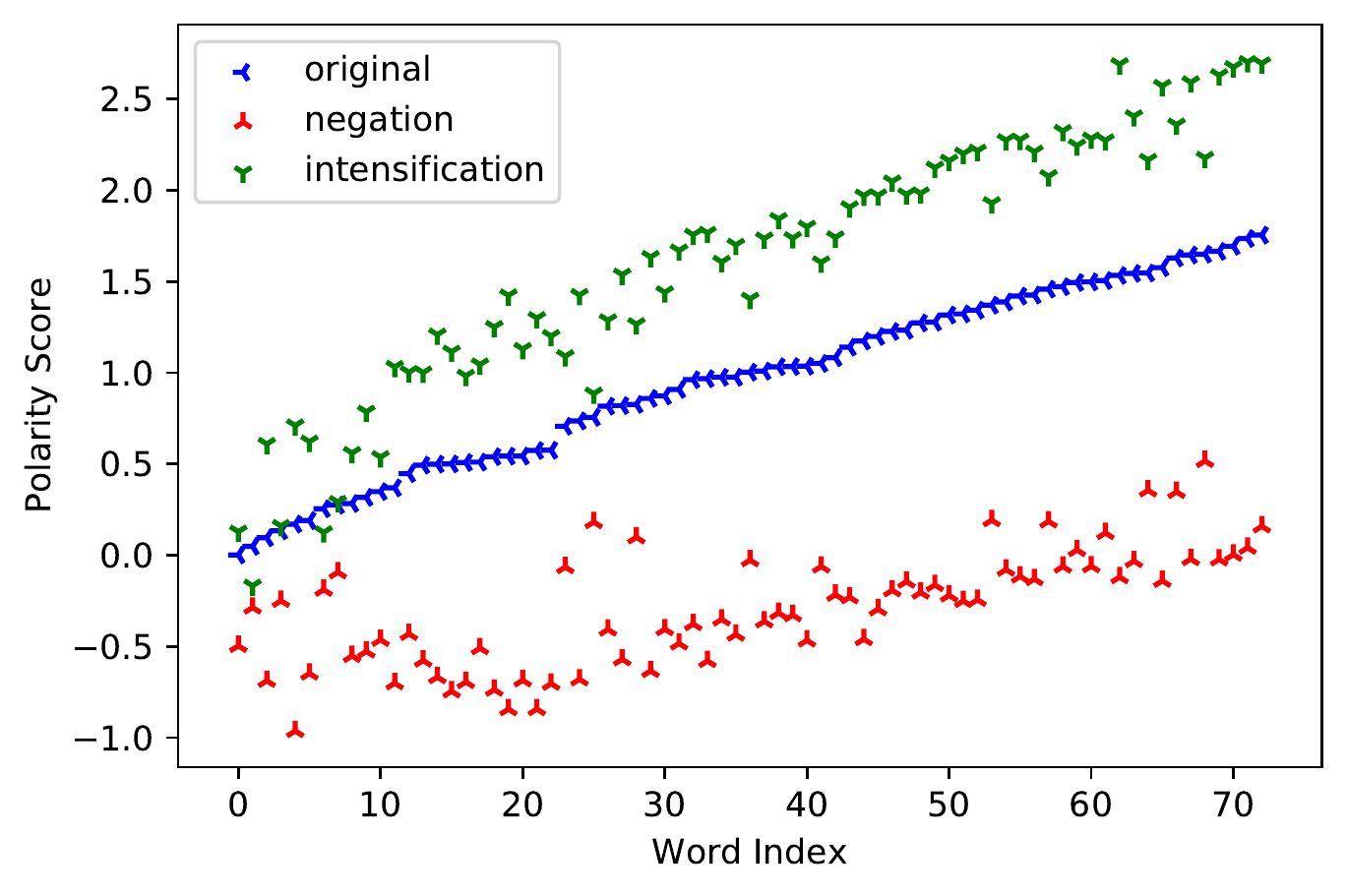}
      \vspace{-3mm}
      \caption{GRU\textsubscript{context}, positive adjectives}
      \label{fig:sst5_int_neg_overall}
    \end{subfigure}
    \begin{subfigure}{0.43\textwidth}
      \centering
      \includegraphics[scale=0.4]{figs/intensification/gru_linear_sst5_negation_intensification_negative.pdf}
      \vspace{-3mm}
      \caption{GRU\textsubscript{context}, negative adjectives}
      \label{fig:sst5_int_neg_phrase}
    \end{subfigure}
    \vspace{-3mm}
    \caption{\color{black}Context polarity scores for positive adjectives (a) and negative adjectives (b) along with their corresponding negation and intensification bigrams from SST-5.  }
    \label{fig:gru_negation_intensification_example2}
    \vspace{-3mm}
\end{figure}

Figure \ref{fig:lstm_negate_adjectives} shows that the $n$-gram representations from the LSTM model together with its corresponding MVMA-L and MVM-L models can also capture negation on the extracted adjectives from SST-2. However, VA-EW fails to capture the negation phenomenon for the negative adjectives, which may be explained by that: the $n$-gram representation of VA-EW solely involves the current token, thus being less expressive compared to the one from models such as MVMA-L and MVMA-G.
Moreover, the MVMA-G model can also capture the negation and intensification phenomena on SST-5 as shown in 
Figure \ref{fig:gru_negation_intensification_example2}. The intensification token will generally strengthen the polarity of an adjective while the negation token will generally weaken the polarity of it.

\begin{figure}
\centering
\vspace{-0mm}
    \begin{subfigure}{0.24\textwidth}
      \centering
      \includegraphics[scale=0.22]{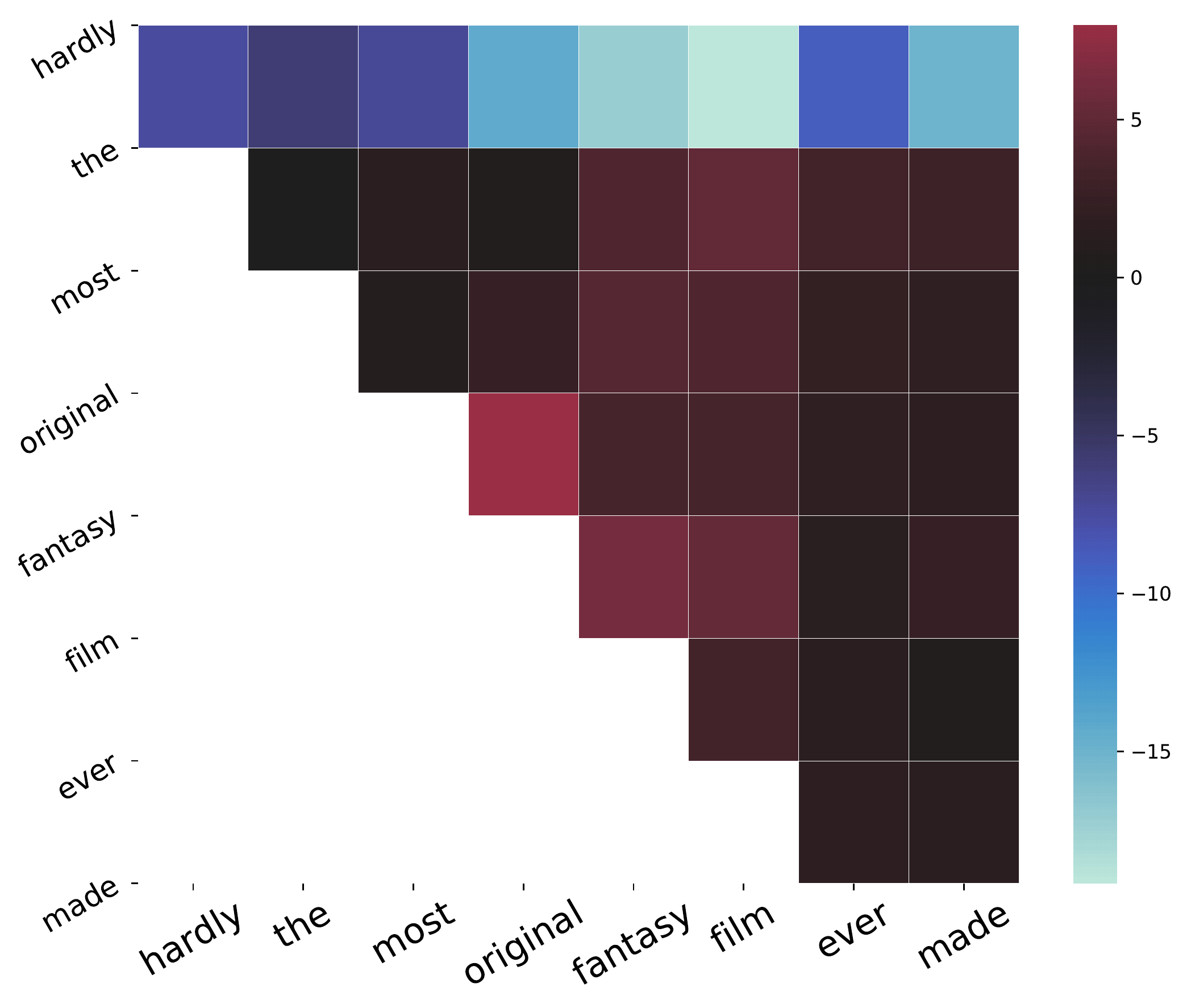}
      \vspace{-4mm}
      \caption{}
      \label{fig:sst2_heatmap1}
    \end{subfigure}
    \begin{subfigure}{0.24\textwidth}
      \centering
      \includegraphics[scale=0.22]{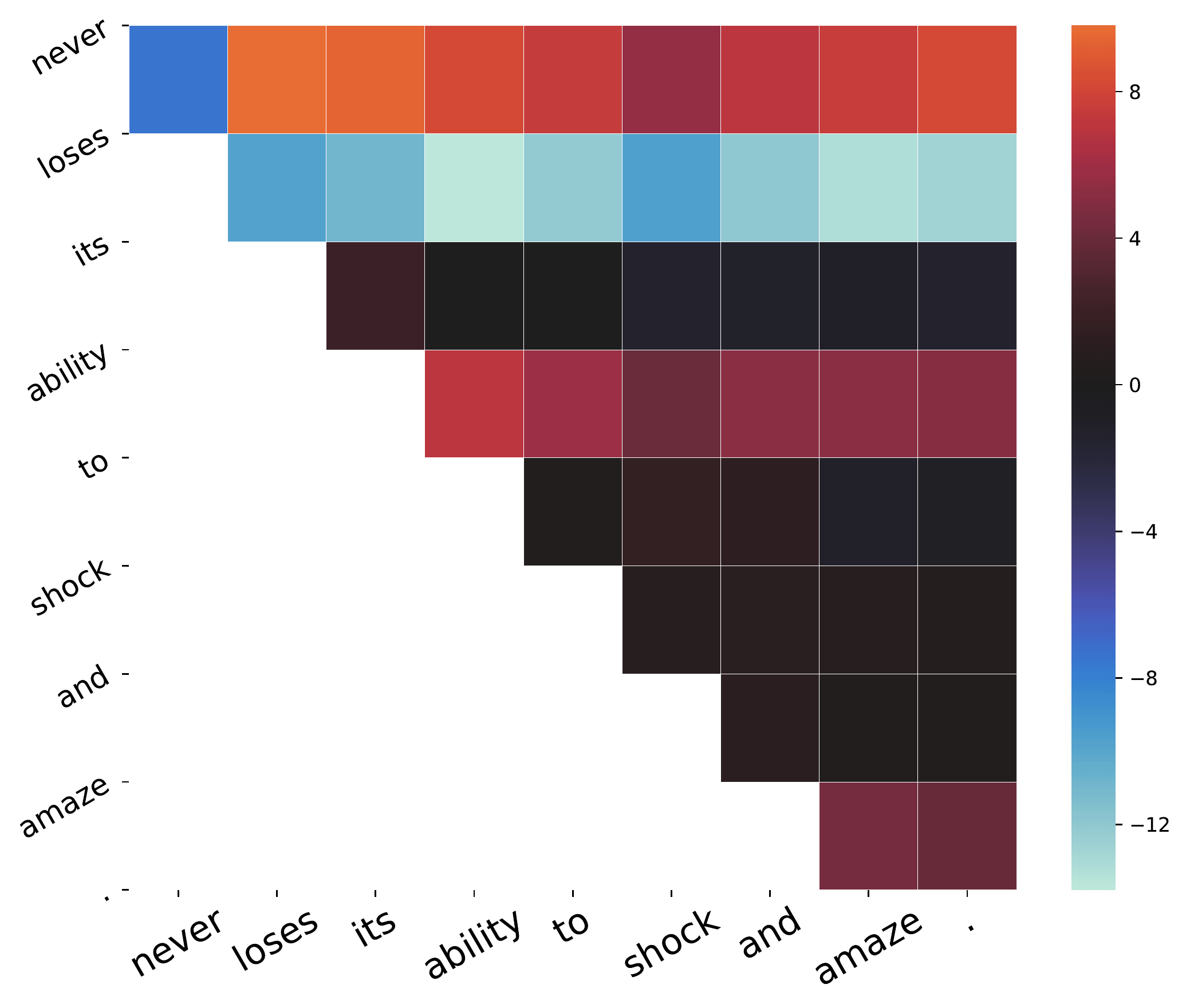}
      \vspace{-4mm}
      \caption{}
      \label{fig:sst2_heatmap2}
    \end{subfigure}
    \vspace{-3mm}
    \caption{Polarity scores for $n$-gram representations within two example sentences. SST-2, MVMA-G.}
    \label{fig:sst_ngram_heatmap}
    \vspace{-4mm}
\end{figure}

We also visualized the polarity score of each $n$-gram within a sentence.
Two examples are shown in Figures \ref{fig:sst2_heatmap1} and \ref{fig:sst2_heatmap2}, where a warmer color indicates a higher polarity score (i.e., the $n$-gram is more positive).
For example, ``{\em never}'' itself has a remarkably negative polarity score while ``{\em loses}'' has a remarkably positive one.
Consequently, the $n$-grams starting from ``{\em never}'' (while ending with another word) generally have positive polarity scores.
Such visualization results show that our identified representations defined over the linguistic units as captured by RNNs can be highly interpretable.

\subsection{First-order Approximation}
\label{subsect:approx_rnn_error}
To examine how well the recurrence relation in Equation \ref{approx:hidden_recurrence_relation} can approximate the standard RNNs, we followed the method in the work of \citet{pmlr-v119-maheswaranathan20a} and compared the hidden state of the standard RNNs ($\vh_t = RNN(\vx_t, \vh_{t-1})$) at each time step to the corresponding context representations ($\hat{\vh}_t = \vg(\vx_t)+\mA(\vx_t)\vh_{t-1}$). The error at each time step is defined as
\begin{align}
    ||\vh_{t}-\hat{\vh}_t||_2/||\vh_{t}||_2.
\end{align}
We used the current standard hidden state to predict the next hidden state and the context representations on the SST-2 test set.

We noticed that the weight decaying coefficient has a remarkable impact on the error. Specifically, a larger coefficient can result in smaller errors.
When the coefficient is $1e-5$, the average errors on the Elman, GRU, and LSTM models were 26.2\%, 21.7\% and 46.6\% and respectively. When the coefficient is $3e-4$ the the average errors dropped to 17.1\%, 15.1\%, and 33.3\% respectively.
Note that since this is the single step error, the accumulated errors across many times steps can be large, particularly for LSTM, and thus the first-order approximation cannot fully replace standard RNNs. Despite this, the resulting context and $n$-gram representations can help us understand how RNNs process contextual information such as $n$-gram features.

\section{T-sne Visualization}
\label{sec:inter_ngrams}

\begin{figure}[t!]
\centering
\vspace{-0mm}
    \begin{subfigure}{0.24\textwidth}
      \centering
      \vspace{5.5mm}
      \includegraphics[scale=0.172]{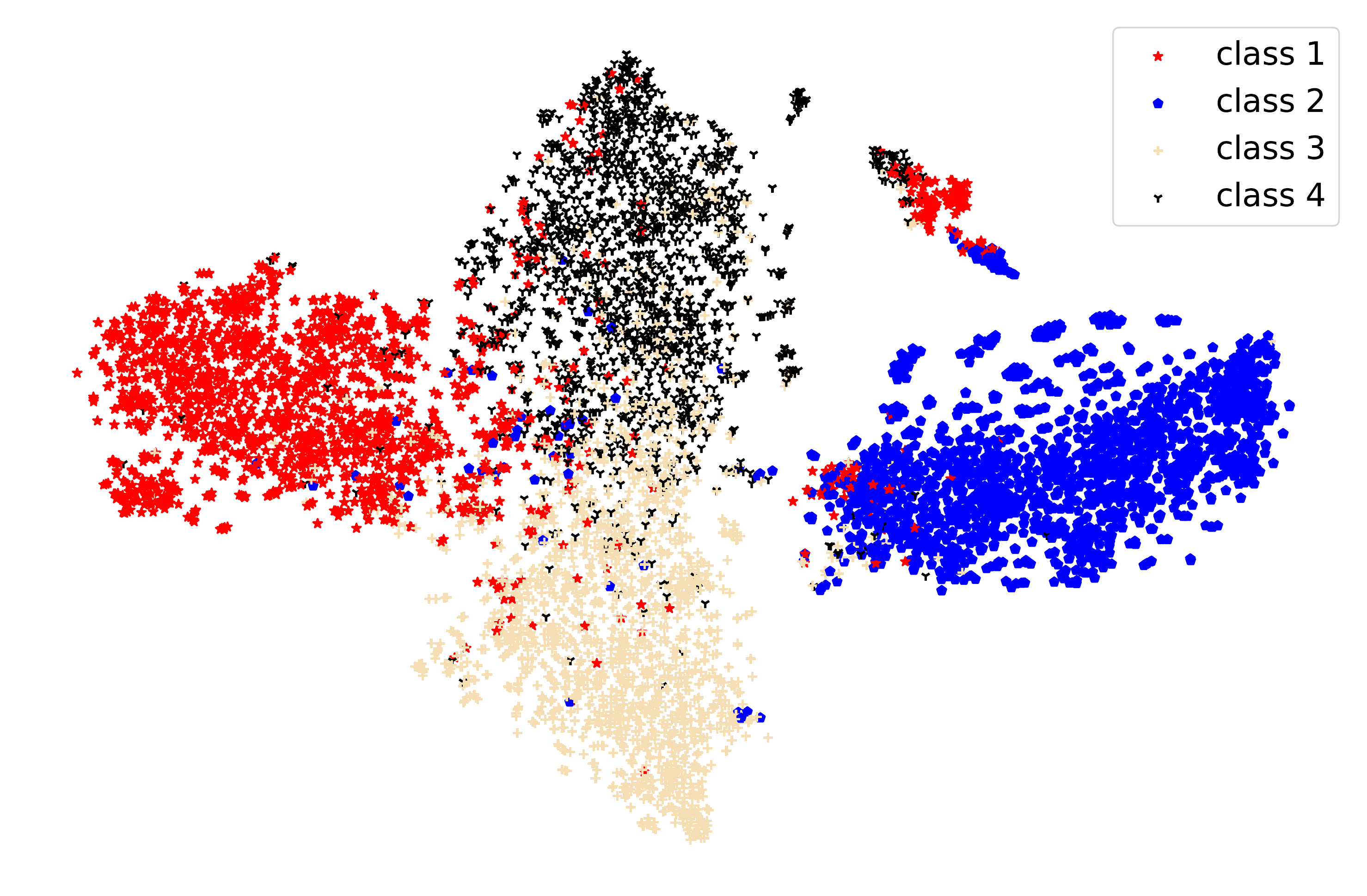}
      \vspace{-4mm}
      \caption{MVMA-G}
      \label{fig:agnews_tsne_mvma}
    \end{subfigure}
    \begin{subfigure}{0.24\textwidth}
      \centering
      \vspace{5.5mm}
      \includegraphics[scale=0.172]{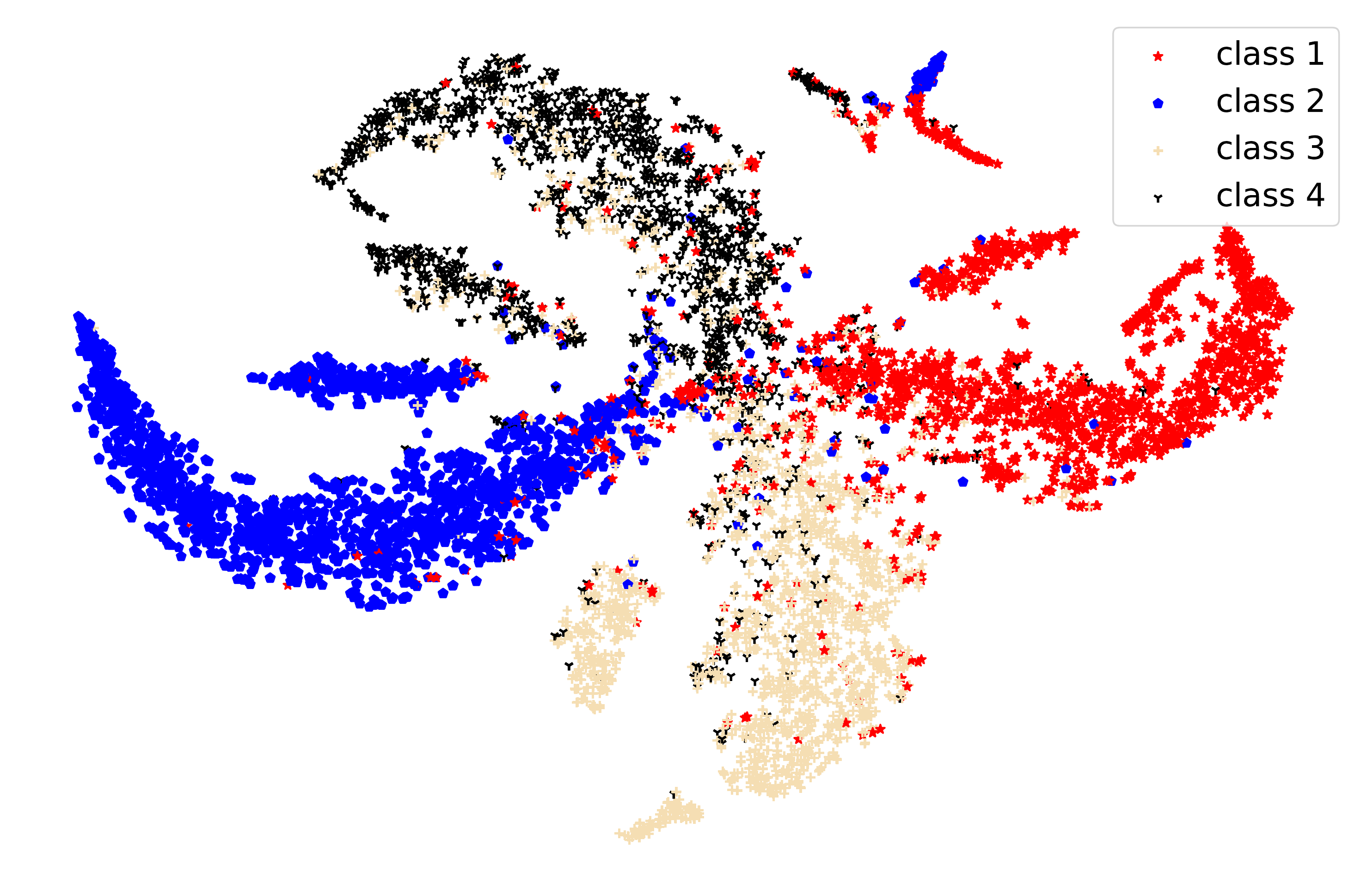}
      \vspace{-4mm}
      \caption{MVM-G}
      \label{fig:agnews_tsne_mvm}
    \end{subfigure}
    \vspace{-3mm}
    \caption{(a) and (b): T-sne visualization of the context representation for phrases (<30 tokens) from the AG-news dataset with four topics.}
    \label{fig:agnews_overall_ngram_component_tsne}
    \vspace{-4mm}
\end{figure}

We visualized the context representations \textcolor{black}{from the MVMA-G model} using t-sne \citep{JMLR:v9:vandermaaten08a}, which provides us with an intuitive understanding on the efficacy of our identified representations.
We automatically extracted 2,188 phrases with less than 30 tokens from  AG-news with 4 topics\footnote{Although SST-5 has 5 lables, most of its phrases are neutral, we therefore did not use this dataset for visualization.} and projected their context representations to a two-dimension space.  
Figures \ref{fig:agnews_tsne_mvma} and \ref{fig:agnews_tsne_mvm} show there exist four major clusters corresponding to the four topics, indicating those representations can generally learn the topic information and explain the differences. 
{\color{black}Similar to the previous analysis, the MVM-G model is able to learn the topic information with the $n$-gram representations.}

\section{Results on Transformer}
We have also run the Transformer model on the sentiment analysis datasets and the results are listed in Table \ref{tab:transformer_performance}.

\begin{table}[t!]
\scalebox{0.65}{
\begin{tabular}{llllll}
\toprule
\multicolumn{2}{c}{\textbf{SST-2}}            & \multicolumn{2}{c}{\textbf{AG-news}}          & \multicolumn{2}{c}{\textbf{IMDB}}    \\
\midrule
dev                 & test                  & dev                 & test                  & dev        & test                  \\
{83.4$\pm$0.4} & {82.0$\pm$0.1} & {90.9$\pm$0.5} & {90.5$\pm$0.4} & 88.4$\pm$0.2 & {88.1$\pm$0.2}
\\
\bottomrule
\end{tabular}
}
\caption{Accuracy on sentiment analysis tasks. Transformer}
\label{tab:transformer_performance}
\end{table}

\section{Implementation Details}
\subsection{Sentiment Analysis}
\paragraph{Settings}
For the SST-2, AG-news, and IMDB datasets, we used the cross-entropy as the loss function to train the models. Embeddings were randomly initialized and trainable during training.
For the SST-5 dataset, we treated the classification as a regression problem as the labels are polarity intensity. The mean-squared error was used as the loss function during training. 
Note that we initialized embeddings with pre-trained GloVe \cite{pennington2014glove} and fixed them during training on SST-5 for the analysis of both the negation and intensification phenomena.

Furthermore, for the MM model, each token was represented as a matrix and the matrix size was set as 32$\times$32. For the other models, the embedding and hidden sizes were both set as 300.

\paragraph{Polarity Adjectives}
We automatically extracted adjectives with polarity (examples shown in Table \ref{tab:selected_adjectives_polarity}) from SST-2 in two steps.
In the first step, following the method of \citet{sun-lu-2020-understanding2}, we calculated a frequency ratio for each token (in the vocabulary) between the frequencies of the token seen in the
positive and negative instances respectively. If a token has a frequency ratio either larger than 3 or less than 1/3, it will be extracted as an \textit{positive token} or an \textit{negative token}.
In the second step, we used the \textit{textblob} package \footnote{https://textblob.readthedocs.io/en/dev/} to detect positive and negative adjectives from those positive tokens and negative tokens respectively. 
\begin{table*}[h]
\centering
\scalebox{0.8}{
\begin{tabular}{cll}
\hline
\textbf{Type} & \multicolumn{1}{c}{\textbf{Adjectives}}                                                          & \multicolumn{1}{c}{\textbf{Size}} \\
\hline
Pos     & \begin{tabular}[c]{@{}l@{}}outstanding, ecological, inventive, comfortable, nice, authentic, spontaneous, sympathetic, lovable,\\  unadulterated, controversial, suitable, grand, happy, enthusiastic, adventurous, successful, noble,\\  true, detailed, sophisticated, sensational, exotic, fantastic, remarkable, impressive, charismatic,\\  good, effective, rich, popular, unforgettable, famous, comical, energetic, ingenious, extraordinary, ...\end{tabular} & 73                                \\
\hline
Neg      & \begin{tabular}[c]{@{}l@{}}bad, tedious, miserable, psychotic, didactic, inexplicable, feeble, sloppy, disastrous, stupid,\\ amateurish, false, cynical, farcical, terrible, unhappy, horrible, atrocious, idiotic, wrong, pathetic,\\ angry, uninspired, vicious, unfocused, unnecessary, artificial, troubled, questionable, arduous,\\ stereotypical, ...\end{tabular}                                                                           & 47
\\
\bottomrule
\end{tabular}}
\caption{Examples of the extracted adjectives from the SST-2 dataset. ``Pos'' refers to \textit{positive} adjectives and ``Neg'' refers to \textit{negative} adjectives.}
\label{tab:selected_adjectives_polarity}
\vspace{-4mm}
\end{table*}

\subsection{Relation Classification}
Following the work of \citet{gupta-schutze-2018-lisa}, we examined the RNN, baseline, MVMA and MVM models on SemEval 2010 Task 8 \citep{hendrickx-etal-2010-semeval} which has 9 directed relationships and an undirected \textit{other} type. We used the final hidden states of the standard RNNs (or context representations of the MVMA, MVM and baseline models) as the instance representations for classification. The cross-entropy loss was employed during training.

\subsection{Named Entity Recognition}
At each time step, we concatenated the context representations (or hidden states) from both directions in a bidirectional model, fed them to a projection layer and then to a linear CRF layer. More details about the architecture can be referred to the biLSTM-CRF model in the work of \citet{lample-etal-2016-neural}. We also referred to the code at \href{https://github.com/allanj/pytorch_neural_crf}{https://github.com/allanj/pytorch\_neural\_crf} for the implementation of the linear CRF layer.

CoNLL-2003 contains four types of entities: persons (PER),
organizations (ORG), locations (LOC) and miscellaneous names (MISC).
The original dataset was labeled with the BIO (Beginning-Inside-Outside) format. For example, ``United Arab Emirates'' are labeled as ``B-LOC I-LOC I-LOC''.
We transformed the tags into the IOBES format where two prefixes ``E-'' and ``S-'' are added. Specifically, ``E-'' is used to label the last token of an entity span. The ``S-'' prefix is used for a single-token span. For example, ``United Arab Emirates'' are labeled as ``B-LOC I-LOC E-LOC'' in this format.
There are 20 categories of tags in total including the starting, ending and padding tags.
We trained the models to predict each entity.


The embedding size and hidden size were set to 300 and 200 respectively. The SGD optimizer was used to learn parameters.

\subsection{Language Modeling}
The embedding size and hidden size were both 512 for PTB and Wiki2, and 256 and 512 respectively for Wiki103. 
The cross-entropy loss was used during training. 
For PTB and Wiki2, the output of the final fully-connected layer was fed to a \textit{softmax} function while the {Adaptive softmax} \citep{joulin2017efficient} was used for Wiki103 (because of its large vocabulary size). We only considered the word-level models.
We trained each model for 50 epochs, chose the model which had the best performance on the development set as the final model and evaluated the final model on the test set. 

\section{Jacobian matrix of LSTM}
Unlike GRU and Elman RNN, LSTM has a memory cell apart from a hidden state. Here, we describe how to get their Jacobian matrices.
An LSTM cell can be written as
 \begin{align}
  \vi_t &= \sigma (\mW_{ii} \vx_t + \mW_{hi} \vh_{t-1}),\nonumber \\
  \vf_t &= \sigma (\mW_{if} \vx_t + \mW_{hf} \vh_{t-1}),\nonumber\\
  \vo_t &= \sigma (\mW_{io} \vx_t + \mW_{ho} \vh_{t-1}),\\
  \vc^m_t &= \tanh (\mW_{ic} \vx_t +  \mW_{hc} \vh_{t-1}),\nonumber\\
  \vc_t &= \!  \vf_t \odot \vc_{t-1} \!+\!   \vi_t \odot \vc^m_t,
  \vh_t \!=\! \vo_t \odot \tanh(\vc_t)\nonumber,
  \label{eq:lstm_standard_cell}
 \end{align}    
 where 
 $\vi_t$, $ \vf_t$, $ \vo_t \in \sR^d$ are the input gate, forget gate and output gate respectively. $\vc^m_t \in \sR^d$ is the new memory, and $\vc_t$ is the final memory. 
 
Let us expand the memory state and hidden state at time step $t$ as
\begin{equation}
    \begin{aligned}
      \vc_t &= \vg_c(\vx_t) + B(\vx_t)\vc_{t-1} \\
      &+ D(\vx_t)\vh_{t-1} + \vo_c(\vc_{t-1}, \vh_{t-1}),\\
      \vh_t &= \vg_h(\vx_t) + E(\vx_t)\vc_{t-1} \\
      &+ F(\vx_t)\vh_{t-1} + \vo_h(\vc_{t-1}, \vh_{t-1}),\\
    \end{aligned}
\end{equation}
where $B$, $D$, $E$ and $F \in \sR^{d \times d}$ are all Jacobian matrices. $\vo_c(\vh_{t-1})$ and $\vo_h(\vh_{t-1})$ are remainder terms of the Taylor expansion.

We concatenate the memory state and hidden state and view the concatenation as an ``extended hidden state''. The context representation for the ``extended hidden state'' at time step $t$ (assuming of zero vectors as initial states) will be written as:
\begin{equation}
\begin{aligned}
    \begin{bmatrix}
    \hat{\vc}_t\\
    \hat{\vh}_t
    \end{bmatrix}
    \!=\! \sum_{i=1}^{t} 
    \begin{bmatrix}
        \vv^c_{i:t}\\
    \vv^h_{i:t}
    \end{bmatrix}
    \!=\!
    \sum_{i=1}^{t} \left[ \prod^{i+1}_{k=t} A(x_k)\right]
    \begin{bmatrix}
         g_c(\vx_i)\\
    g_h(\vx_i)
    \end{bmatrix},
\end{aligned}    
\label{eq:lstm_hidden_state_unroll5}
\end{equation}
where $\hat{\vc}_t$ and $\hat{\vh}_t$ refer to the context representations corresponding to the memory state and hidden state respectively. $g_c$, $g_h \in \sR^d$, and $A \in \sR^{2d \times 2d}$ are all functions of inputs.
$A(x_k)$ contains many interaction terms resulting from the gating mechanism, which may result in a strong expressive power.
As the hidden state $\vh_t$ is commonly used for downstream tasks, we will only consider $\vv^h_{i:t}$ as the $n$-gram representation on our tasks, and the context representation will be $ \sum_{i=1}^{t} \vv^h_{i:t}$.